\documentclass[onecolumn,lettersize,journal]{IEEEtran}
\usepackage[caption=false,font=normalsize,labelfont=sf,textfont=sf]{subfig}
\def\BibTeX{{\rm B\kern-.05em{\sc i\kern-.025em b}\kern-.08em
		T\kern-.1667em\lower.7ex\hbox{E}\kern-.125emX}}
\usepackage{balance}
\usepackage[utf8]{inputenc}
\usepackage{geometry}
\usepackage{dirtytalk}
\usepackage{graphicx}
\usepackage{wrapfig}
\usepackage{lipsum}
\usepackage{booktabs}
\usepackage{listings}
\usepackage[lined,linesnumbered,ruled,noend]{algorithm2e}
\usepackage{algpseudocode}
\usepackage{array}
\usepackage{amsmath}
\usepackage{amsfonts}
\usepackage{url}
\usepackage{verbatim}
\usepackage{stfloats}
\usepackage{textcomp}
\usepackage{tocbibind}
\usepackage{pgfplots}
\usepackage{pdfpages}
\pgfplotsset{compat=1.16}
\usepackage{tikz}
\usepackage{cite}
\usepackage{xcolor}
\usepackage{colortbl}
\usepackage{multirow}
\usetikzlibrary{patterns}

\DeclareMathAlphabet{\mathpzc}{OT1}{pzc}{m}{it}

\newtheorem{lemma}{Lemma}
\newtheorem{theorem}{Theorem}

\newtheorem{prooft}{Proof}
\newtheorem{assumption}{Assumption}

\begin{document}
	
	\title{FedSat: A Statistical Aggregation Approach for Class Imbalanced Clients in Federated Learning}
	
	\author{Sujit~Chowdhury, Raju~Halder
		\IEEEcompsocitemizethanks{\IEEEcompsocthanksitem Sujit~Chowdhury and Raju~Halder are with the Department of Computer Science and Engineering, Indian Institute of Technology Patna, India, 801106. \protect
			( E-mail: \{sujit\_2021cs35, halder\}@iitp.ac.in )}
		\thanks{This work has been submitted to the IEEE for possible publication. Copyright may be transferred without notice, after which this version may no longer be accessible.}}
	
	\markboth{Journal of \LaTeX\ Class Files,~Vol.~14, No.~8, August~2021}%
	{Shell \MakeLowercase{\textit{et al.}}: A Sample Article Using IEEEtran.cls for IEEE Journals}
		
	\maketitle
	
	\begin{abstract}
		Federated learning (FL) has emerged as a promising paradigm for privacy-preserving distributed machine learning, but faces challenges with heterogeneous data distributions across clients. This paper presents \textsf{FedSat}, a novel FL approach specifically designed to simultaneously handle three forms of data heterogeneity, namely \emph{label skewness}, \emph{missing classes}, and \emph{quantity skewness}, by proposing a \emph{prediction-sensitive loss function} and a \emph{prioritized-class based weighted aggregation scheme}. While the prediction-sensitive loss function enhances model performance on minority classes, the prioritized-class based weighted aggregation scheme ensures client contributions are weighted based on both statistical significance and performance on critical classes. Extensive experiments across diverse data-heterogeneity settings demonstrate that \textsf{FedSat} significantly outperforms state-of-the-art baselines, with an average improvement of 1.8\% over the second-best method and 19.87\% over the weakest-performing baseline. The approach also demonstrates faster convergence compared to existing methods. These results highlight \textsf{FedSat}'s effectiveness in addressing the challenges of heterogeneous federated learning and its potential for real-world applications.
	\end{abstract}
	

	\section{Introduction}
	\label{sec:Introduction}
	\IEEEPARstart{F}{ederated} Learning (FL) \cite{fedavg} has emerged as a promising paradigm for training machine learning models across decentralized edge devices, enabling privacy-preserving and efficient model updates without the need to centralize sensitive data. However, the effectiveness of FL is often challenged by various factors, including non-independent and identically distributed (non-IID) datasets, varying network conditions, and heterogeneous devices among clients. This yields biases in model training, leading to sub-optimal performance \cite{non_iid_exp_study, FedLC}. 
	
	Since its inception in 2017, a significant research effort has been observed in the literature to address the above-mentioned challenges. The authors in~\cite{Li2020On, yang2021achieving, TaoLin2020, fedavgM} attempted to improve the performance of FL models by regulating local epochs or server-side updates in presence of non-IID datasets. However, they experience a slow and sometimes unstable convergence due to the diverse local objectives arising from heterogeneous data distributions, commonly known as client drift, as reported in~\cite{scaffold, FedLADA, CCVR}. Similarly, the regularization techniques adopted in~\cite{fedprox, fedDyn, fedNova, scaffold, DANE, fedDANE} often face difficulty due to the skewness in labels, features, and quantity of the data across various clients, leading to a slow and unstable convergence. Furthermore, it is observed that most of the existing approaches employed simple aggregation schemes~\cite{fedavg, scaffold, fedprox, fedDyn, FedLADA, FedLC}, which are unable to deal with extreme heterogeneous settings where clients' models differ from each other significantly due to their local data.
	
	To showcase the impact of heterogeneous data settings in FL, the authors in~\cite{non_iid_exp_study} developed a benchmark with comprehensive non-IID configurations, including label skew, features skew, and quantity skew. The empirical study revealed that none of the existing state-of-the-art FL approaches consistently outperform others under all different configurations. This finding inspires researchers to develop specialized algorithms tailored to specific non-IID configurations for further enhancing the performance of the global model. For instance, the authors in~\cite{FedLC} attempted to mitigate label skewness by utilizing logits calibration which handles missing classes during the local training. Similarly, \cite{FedBN} aimed at addressing the feature distribution skewness by utilizing batch normalization techniques. On the other hand, FedCorr~\cite {FedCorr} is introduced to address data heterogeneity concerning both local label quality and label skewness by computing a Gaussian Mixture Model (GMM) based on the cumulative Local Intrinsic Dimensionality (LID) scores from all clients. 
	
	In summary, existing approaches in the literature either focus on specific aspects of data skewness or require complex data correction models, which can be difficult to generalize across diverse FL environments. Therefore, creating an effective FL framework that works well in most situations remains a challenging and open problem in the field of Federated Learning.
	
	To this aim, this paper presents \textsf{FedSat}, a novel federated learning approach specifically designed to simultaneously handle three forms of data heterogeneity, namely \emph{label skewness}, \emph{missing classes}, and \emph{quantity skewness}. In particular, \textsf{FedSat} introduces two key components: \textbf{\emph{prediction-sensitive loss function}} and \textbf{\emph{prioritized-class based weighted aggregation scheme}}. The proposed prediction-sensitive loss function significantly enhances model performance by assigning higher misclassification penalties to minority classes, thereby effectively managing class imbalance. Meanwhile, the prioritized-class based weighted aggregation scheme ensures that clients' contributions to the global model are weighted based on both the statistical significance of their locally trained models and their performance on critical classes. This combined approach enhances the robustness and accuracy of the global model, particularly when faced with extreme non-IID data heterogeneity.
	
	\textsf{FedSat} operates in three key stages: (1) the server randomly selects a subset of nodes as clients for performing local model training, employing the novel prediction-sensitive loss function; (2) for each client's parameters, the server forms a worker-set comprising randomly sampled non-client nodes and the client itself to analyze statistical performance of the trained parameters. This statistical analysis provides weight scores which balance the impact of both label and quantity skews by identifying underperforming classes; and (3) finally, the server utilizes the updated local parameters and the computed weight scores to generate more generalized and robust global parameters, effectively mitigating various form of data heterogeneity altogether.

	We conduct an extensive set of experiments across diverse data-heterogeneity settings, showcasing a substantial  performance enhancement with the \textsf{FedSat} compared to the baselines~\cite{chen2023elastic, fedavg, fedavgM, fedDyn, FedLC, fedprox, FedLADA, scaffold}, even in case of extreme non-convex data settings. 
	
	\subsection{Contributions}
	\begin{itemize}
		\item We introduce \textsf{FedSat}, a novel federated learning approach designed to achieve robust and generalized global models in highly heterogeneous data settings.
		\item We propose a \emph{prediction-sensitive loss function} to address the effect of label skewness and missing classes during local training.
		\item  We propose \emph{prioritized-class based weighted aggregation scheme} which, unlike traditional aggregation schemes, leverages class-wise statistical insights to enhance the performance of critical classes and generates a robust global model.
		\item We perform an extensive experiments across diverse data-heterogeneity settings, which demonstrate a significant performance improvement of \textsf{FedSat} compared to the baselines, while also ensuring the robustness of the global model.
	\end{itemize}
	
	\noindent The rest of the paper is organized as follows: Section \ref{sec:Related Works} presents an overview of the related work. We present the details of \textsf{FedSat} in Section \ref{sec:Proposed work}. The convergence analysis of \textsf{FedSat} is presented in Section \ref{sec:convergence proof}. Section \ref{sec:implementation and evaluation} provides the prototype implementation and detailed experimental evaluation. Finally, Section \ref{sec:conclusion} concludes the paper.

	\section{Related Work}
	\label{sec:Related Works}
	To address the issue of users data privacy and communication bottleneck in distributed learning, McMahan et al. first introduced the federated learning algorithm FedAvg~\cite{fedavg} in 2017. This approach allows decentralized devices to train models locally, followed by the aggregation of their trained parameters on a central server. Since then, significant advancements in federated learning approaches have been observed in the literature, primarily addressing various challenges of FL settings, such as high communication costs, data heterogeneity, model divergence, client drift, and label skewness. To speed up the convergence of FedAvg in heterogeneous settings~\cite{Li2020On, yang2021achieving, TaoLin2020, fedavgM}, a series of SGD-based approaches have been proposed, especially for large-scale training in Federated Learning. In~\cite{fedavgM} the authors proposed FedAvgM, aiming to address the slow convergence of FedAvg, particularly in scenarios with non-IID data across diverse clients, by adding a momentum in the global model update process on the server. However, due to the integration of momentum, careful tuning of hyper-parameters is required to converge the global model. To further tackle heterogeneity in FL, a list of approaches~\cite{fedprox, fedDyn, fedDc, AdaBest, FedPage, 10040221} applied regularization method. FedProx~\cite{fedprox} introduced a temperature parameter and a proximal regularization term into the optimization function to control the model drift between client and server, especially mitigating the client straggler issue in a heterogeneous system. However, because of the proximity term, local updates shift towards the last global update, which does not assure alignment between global and local optima. To this aim, FedDyn~\cite{fedDyn} introduced a dynamic regularization term in the local loss function, aligning local updates with global model parameters. This approach mitigates client drift arising from local overfitting. Recently, authors in~\cite{FedLADA} proposed a momentum-based algorithm, FedLADA, which achieves linear speedup in convergence by mitigating client drift issues due to local overfitting via utilization of the global gradient descent and locally adaptive amended optimizer. Likewise, some studies~\cite{DANE, scaffold, ProxSkip, fedDyn, chen2023elastic } use client-specific control variates, a variance reduction technique, to control the update direction. SCAFFOLD~\cite{scaffold}, which is an improvement of DANE~\cite{DANE}, employs additional server-specific control variates to address the client drift in its local updates. Chen et al. \cite{chen2023elastic} proposed Elastic aggregation, which adaptively interpolates client models based on parameter sensitivity to prevent server model drift and improve convergence in federated learning with non-IID data.
	
	Another line of work develops FL algorithms based on the feature representation of the training model~\cite{CollinsHMS22, CCVR, MOON, yu2022tct}. MOON~\cite{MOON} addresses non-IID data distributions among clients by utilizing similarity between model representations to correct the local model updates of individual clients. CCVR~\cite{CCVR} introduced the classifier regularization and calibration method to improve FL performance by fine-tuning the classifier using virtual representations sampled from an approximated Gaussian mixture model.
	
	An attempt to tackle class skewness in the datasets across clients is reported in~\cite{FedCorr, FedLC}. FedCorr~\cite{FedCorr} introduced an FL framework for handling label noise while protecting data privacy. It comprises three stages: identifying and correcting noisy data, fine-tuning the model using cleaner data, and standard training with all corrected data. To improve the performance of FL approaches in label-skewed dataset distribution across clients, ~\cite{FedLC} introduced FedLC, which adjusts logits before computing cross-entropy loss based on the probability of each class occurrence.
	
	\begin{figure}[t]
		\centering
		\includegraphics[width=\columnwidth]{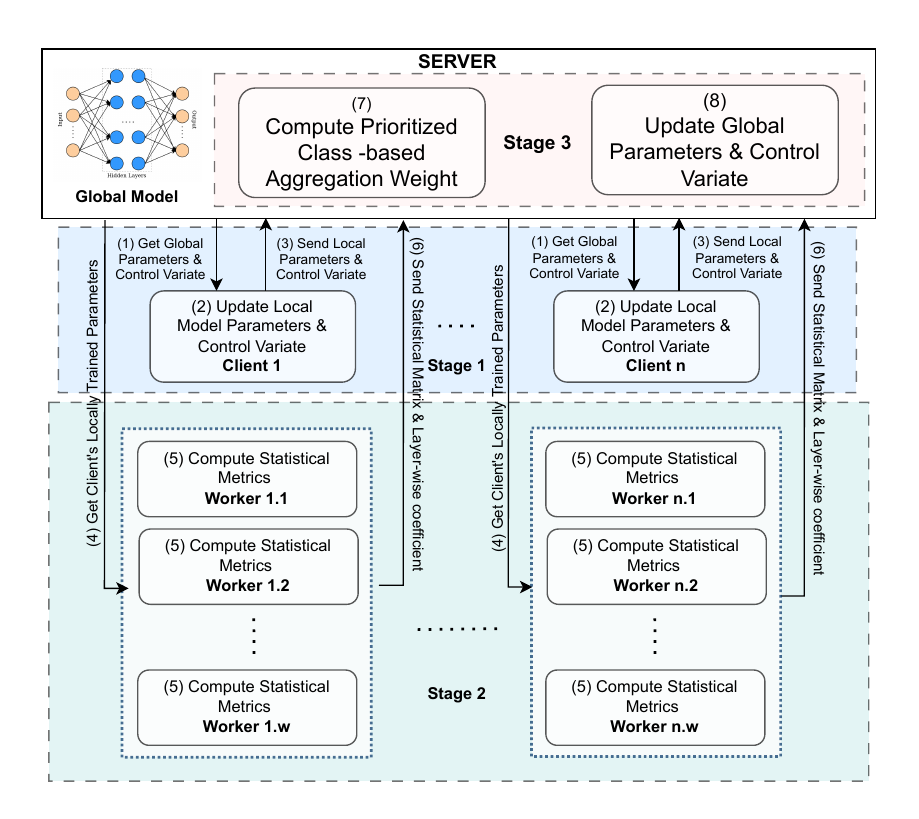}
		\caption{\textsf{FedSat} Framework Architecture.}
		\label{fig:framework}
	\end{figure}
	
	\section{\textsf{FedSat}: Proposed FL Approach}
	\label{sec:Proposed work}

	\textsf{FedSat} introduces two modular components that enhance federated learning algorithms to better handle data heterogeneity:
	(1) prediction-sensitive loss function for client's local training, and (2) prioritized-class based weighted aggregation scheme for aggregating local updates. Figure~\ref{fig:framework} depicts the architectural overview of \textsf{Fedsat}. It involves a set of nodes acting as clients and workers in a particular round. The clients are responsible for locally training the global model parameters and sending the updated parameters back to the server (Steps 2 and 3), whereas workers are responsible for computing class-wise statistical performance metrics of their associated client's parameters (Step 5). The server facilitates data communication among the selected clients and workers in each round (Steps 1, 3, 4, and 6), and it generates global model by performing our proposed prioritized-class based weighted aggregation method (Steps 7, 8). It is worthwhile to note that the number of workers employed to evaluate each client's model performance is a trade off between evaluation robustness and communication cost. The more number of workers employed to evaluate clients parameters the more robust and generalize the evaluation score is.

	\subsection{Learning Objectives}
	
	Let $\mathcal{K}$ be the set of clients participated in the system and $m$ be the number of distinct classes present in the datasets available with all the clients. The dataset $\mathcal{D}_k$ associated with the client $k \in \mathcal{K}$ is defined as: $\mathcal{D}_k = \{ d_k^i | i=1 \ldots m \}$, where $d_k^i=\{(x_k^{ij}, l_k^{ij}) | j=1 \ldots n_k^i\}$ is the set of data-samples $x_k^{ij}$ and their associated labels $l_k^{ij}$ in the $i^{th}$ class. Observe that, due to the heterogeneity, the number of data samples in different classes may vary, i.e. $|d_k^i|\ge 0$. 
	
	In \textsf{FedSat}, our primary goal is to achieve a resilient global model $\Theta^*$ through collaborative training of local models across the clients, even in presence of various heterogeneous settings. To formalize, we characterize the optimal global model $\Theta^*$ as follows:
	\begin{equation}
		\Theta^* = arg\underset{\Theta}{min}\sum_{k \in \mathcal{K}} \vartheta_k L(\theta_k)
	\end{equation}
	where 
	\begin{itemize}
		\item[-] $\Theta = \left\{ \theta_k | k \in \mathcal{K} \right\}$ is the set of locally trained parameters of the clients,
		\item[-] $\vartheta_k$ is the aggregation weight assigned to the parameters of the client $k$ using our proposed statistical weight calculation algorithm, and 
		\item[-]$L(\theta_k)$ is the loss function defined for a batch of $\mathpzc{N}$ random samples in $\mathcal{D}_k$ in terms of prediction-sensitive loss $\mathcal{L}_{PS}$ as follows:
		\begin{equation}
			\label{eq:simple_loss_function}
			L(\theta_k) = \frac{1}{\mathpzc{N}} \sum_{r=1}^{\mathpzc{N}} \mathcal{L}_{PS}^r(\theta_k;x_k^{ij},l_k^{ij}), 
		\end{equation}
		where $(x_k^{ij}, l_k^{ij})\in \mathcal{D}_k$
	\end{itemize}
	
	\subsection{Local Training with Prediction-Sensitive Loss}
	
	At the commencement of each federated round \( t \), the server randomly selects a subset \( S_t \subseteq \mathcal{K} \) of nodes as clients and distributes the global model parameters \( \Theta_t \) among them. Each selected client then load the global model parameters to its local model (setting \( \theta_k = \Theta_t \)) and performs local training using our proposed prediction-sensitive loss function defined below:
	
	\begin{equation}
		\label{eq:cost_sensitive_loss_func}
		\mathcal{L}_{PS}(\theta_{t,k}^{e}) = \sum_{i=1}^{|\mathcal{D}_k|} cs_{l_i, \hat{l}_i} \cdot \mathcal{L}(\theta_{t,k}^{e}, l_{i}, \hat{x}_{i})
	\end{equation}
	
	\noindent where \( \mathcal{L} \) is the standard loss function (e.g., cross-entropy loss \cite{cross_entropy_loss}), $\hat{x}_i$ is the logits of sample $x_i$, $\hat{l}_i$ is the predicted class from the logits $\hat{x}_i$, and \( cs_{l_i, \hat{l}_i} \) represents the cost associated with predicting a sample of class \( l_i \) as \( \hat{l}_i \). The value of $cs_{l_i, \hat{l}_i}$ is determined by the following equation:
	
	\begin{equation}
		cs_{l_i,\hat{l}_i} =
		\begin{cases} 
			1 & \text{if } l_i = \hat{l_i}, \\
			\epsilon_1 + \frac{\mathcal{M}_{l_i, \hat{l}_i} - \underset{p,q\in m}{min}(\mathcal{M}_{p,q})}{\underset{p,q\in m}{max}(\mathcal{M}_{p,q}) - \underset{p,q\in m}{min}(\mathcal{M}_{p,q})} (\epsilon_2 - \epsilon_1) & \text{if } l_i \neq \hat{l_i}.
		\end{cases}
	\end{equation}
	
	\noindent Here:
	\begin{itemize}
		\item $\mathcal{M}_{l_i, \hat{l}_i}$ represents the number of times the model predicted class $\hat{l_i}$ for samples whose true labels are $l_i$.
		\item \( \epsilon_1 \) and \( \epsilon_2 \) are tunable parameters controlling the range of the penalty for misclassification.
		\item If the prediction is correct (\( l_i = \hat{l}_i \)), the cost is set to 1 indicating there is no penalty.
		\item For incorrect predictions (\( l_i \neq \hat{l}_i \)), the cost is a normalized value derived from the evaluation matrix, scaled between \( \epsilon_1 \) and \( \epsilon_2 \).
	\end{itemize}
	
	Observe that compared to the standard loss function our proposed loss function applies a prediction-driven dynamic loss refinement mechanism. This adaptive mechanism efficiently amplifies the loss based on the distribution of misclassification patterns. Particularly, we introduce a sophisticated cost matrix $C=[cs_{l_i,\hat{l_i}}]_{m\times m}$ that captures and quantifies the model's misclassification tendencies arising from data distribution skewness. By incorporating these learned patterns, the cost matrix dynamically adjusts the loss magnitude, consequently amplifying the gradients step and enhancing the model's learning trajectory.

	During local training, each client \( k \in S_t \) incorporates this prediction-sensitive loss function and updates the local model according to the following Equation for each epoch \( e \) in \( t \)-th round:
	
	\begin{equation}
		\label{eq:local_per_epoch_update}
		\theta_{t,k}^{e+1} = \theta_{t,k}^{e} - \eta_l \left( \frac{\partial \mathcal{L}_{PS}(\theta_{t,k}^{e})}{\partial \theta_{t,k}^{e}} + \gamma(v_t - v_{t,k}) \right)
	\end{equation}
	
	\noindent where \( \eta_l \) is the client's learning rate. This is interesting to note that, unlike existing variance reduction approaches~\cite{scaffold, fedDANE, fedDyn, FedLADA}, each client in \textsf{FedSat} computes global correction term $v_t$ locally without any additional communication overhead between clients and server, as follows:
	\begin{equation}
		\label{eq:global_correction_term}
		v_{t} = \frac{\eta_l}{\mathbf{E}}(\Theta_t - \Theta_{t-1})
	\end{equation}
	where $\mathbf{E}$ is the number of local epochs. Here the regularization term $\gamma(v_t - v_{t,k})$ enhances both convergence and performance through an adaptive correction mechanism with strength $\gamma$ that dynamically refines the optimization trajectory based on the interplay between global correction term $v_t$ and the local correction term $v_{t,k}$.
	
	Upon completion of local updates, each client \( k \) updates its local correction term \( v_{t,k} \), as follows: 
	
	\begin{equation}
		\label{eq:client_drift_estimation}
		v_{t+1,k} = v_{t,k} - v_t + \eta_l \times \frac{\partial \mathcal{L}_{PS}(\theta_{t,k})}{\partial \theta_{t,k}}
	\end{equation}
	
	\noindent Finally, these updated model parameters \( \theta_{t,k} \) are sent back to the server. The server then aggregates them to update the global model parameters for the next round. This process ensures that the updated local model mitigates possible client drift, while the prediction-sensitive matrix \( C \) helps in addressing class imbalance effectively.
	
	\subsection{Learned parameter analysis}
	For each client's learned parameters $\theta_{t,k}$ (where $k  \in S_t$) at round $t$, the server forms a worker set $W_{t,k} \in (\wp(\mathcal{K} - S_t) \cup k)$ which includes a predefined number of randomly selected non-client nodes as well as the client itself. 
	Each worker in $W_{t,k}$ then loads its assigned client parameters into its model and computes the following crucial matrices. Observe that, the optimal number of workers to be chosen depends on the specific application requirements and conditions, establishing a trade-off between communication cost and model robustness. In scenarios with extreme heterogeneity, where clients exhibit highly imbalanced class distributions and sample sizes, selecting a larger set of workers becomes crucial. This ensures fairness in the evaluation of learned parameters and enhances the system's robustness by allowing the model to be assessed on a sufficient number of samples across all classes. Conversely, in communication-constrained systems, the number of workers can be reduced to one, indicating that only the client itself acts as the worker.
	
	\subsubsection{\textbf{Prediction Matrix (\(\hat{Z}\))}}
	\begin{equation}
		\hat{Z}^{w}_{t,k} = \Big[|\hat{\mathcal{Z}}^{w,i}_{t,k}| ~\Big| \, i \in 1 \ldots m ~\Big]
		\label{eq:prediction_mx}
	\end{equation}
	where \(\hat{\mathcal{Z}}^{w,i}_{t,k} = \{ x_k^{ij} \mid i = \text{arg}\underset{y \in 1 \ldots m}{\text{max}} P(Y = y | X = (x_k^{ij}, \theta_k)) \}\) is the set of samples of the worker $w$ which are predicted as class $i$ utilizing $\theta_k$.
	
	\subsubsection{\textbf{True Positive Matrix (\(TP\))}}
	\begin{equation}
		TP^{w}_{t,k} = \Big[|\mathcal{TP}^{w,i}_{t,k}| ~\Big| \, i \in 1 \ldots m ~\Big]
		\label{eq:TP_mx}
	\end{equation}
	where \(\mathcal{TP}^{w,i}_{t,k} = \{ x_k^{ij} \mid l_k^{ij} = \text{arg}\underset{y \in 1 \ldots m}{\text{max}} P(Y = y | X = (x_k^{ij}, \theta_k)) \}\) is the set of samples of the worker $w$ which are predicted correctly as class $i$ utilizing $\theta_k$.
	
	\subsubsection{\textbf{Target Matrix \((T)\)}}
	\begin{equation}
		T^{w}_{t,k} = \Big[|\mathcal{T}^{w,i}_{t,k}|~ \Big| \, i \in 1 \ldots m \Big]
		\label{eq:target_mx}
	\end{equation}
	where \(\mathcal{T}^{w,i}_{t,k} = \{ x_k^{ij} \mid j=1 \ldots n_k^i \}\) is the set of samples of the worker $w$ which are labeled as class $i$.
	
	The worker then sends these results to the server for weight computation and global model update. 
	
	\subsection{Weight computation}
	After obtaining class-specific matrices from all participating workers at round \(t\), the server proceeds to compute three crucial metrics: False Negative Rate (\texttt{FNR}), False Positive Rate (\texttt{FPR}), and Accuracy (\texttt{ACC}). These metrics helps to assess the class-wise performance of the trained parameters' of each client across the chosen workers.
	The \texttt{FNR} of client \( k \) at round \( t \) is determined by the ratio of incorrectly classified negative instances to the total number of true negative instances, defined below:
	
	\begin{align}
		\texttt{FNR}_{t,k} &= \big[\emph{fnr}_{t,k}^{i} \mid i \in 1 \ldots m \big]
		\label{eq:FNR}
	\end{align}
	where
	\begin{align}
		\emph{fnr}_{t,k}^{i} &= \frac{\sum_{w \in W_{t,k}} (|\mathcal{T}^{w,i}_{t,k}| - |\mathcal{TP}^{w,i}_{t,k}|)}{\sum_{w \in W_{t,k}}|\mathcal{TP}^{w,i}_{t,k}|}
	\end{align}
	Similarly, \texttt{FPR} of client \( k \) at round \( t \) is calculated as the ratio of falsely classified positive instances to the total number of true negative instances, computed as follows:
	\begin{align}
		\texttt{FPR}_{t,k} &= \big[\emph{fpr}_{t,k}^{i} \mid i \in 1 \ldots m \big]
		\label{eq:FPR}
	\end{align}
	where
	\begin{align}
		\emph{fpr}_{t,k}^{i} &= \frac{\sum_{w \in W_{t,k}} (\hat{|\mathcal{Z}}_{w,i}^{k}| - |\mathcal{TP}^{w,i}_{t,k}|)}{\sum_{w \in W_{t,k}}(\sum_{r\in 1 \ldots m}|\mathcal{T}^{w,r}_{t,k}| -|\mathcal{T}^{w,i}_{t,k}|)}
	\end{align}
	Finally, \texttt{ACC} of each client \( k \) at round \( t \) is computed by evaluating the ratio of correctly classified instances to the total number of instances defined as:
	\begin{align}
		\texttt{ACC}_{t,k} &= \big[\emph{acc}_{t,k}^i \mid i \in 1 \ldots m \big]
		\label{eq:ACC}
	\end{align}
	where
	\begin{align}
		\emph{acc}_{t,k}^i &= \frac{\sum_{w \in W_{t,k}}\mathcal{TP}^{w,i}_{t,k}}{\sum_{w \in W_{t,k}}\mathcal{T}^{w,i}_{t,k}}
	\end{align}
	
	\noindent Following the above computations, the server assigns a vector of prioritization score (\( \mathcal{E}_{t,k} \)) to each client \( k \) at round $t$, aiming to identify classes that require enhanced performance. This score considers both false negative and false positive rates, normalized to the maximum rates across all classes, computed below:
	
	\begin{equation}
		\mathcal{E}_{t,k} = [  \mathcal{E}_{t,k}^i \mid i \in 1 \ldots m]
	\end{equation}
	where
	\begin{equation}
		\mathcal{E}_{t,k}^i = \alpha \times \frac{\texttt{FNR}_{t,k}[i]}{\underset{r \in 1 \ldots m}{max}\texttt{FNR}_{t,k}[r]} + \beta \times \frac{\texttt{FPR}_{t,k}[i]}{\underset{r \in 1 \ldots m}{max}\texttt{FPR}_{t,k}[r]}
		\label{eq:Ei}
	\end{equation}
	and $\alpha,~\beta$ are the coefficients used to set the importance of the computed metrics. 
	
	Subsequently, the server selects a priority class (\( \mathcal{PC}_t \)) at round $t$ to focus on improving the performance of the most critical class, based on the sum of prioritization scores across all classes, as follows:
	\begin{equation}
		\mathcal{PC}_t = arg\underset{i \in 1 \ldots m}{max} \, \sum_{k \in S_t} \mathcal{E}_{t,k}^i
		\label{eq:PC}
	\end{equation}
	
	Further, a set of client parameters (\( A_t \)) is selected for aggregation based on their corresponding prioritization scores. Clients with scores exceeding a certain threshold relative to the average score across all classes are chosen for weighted aggregation defined as:
	\begin{equation}
		A_t = \Big\{\theta_{t,k} | \sum_{i \in 1 \ldots m}\mathcal{E}_{t,k}^i \le 
		\frac{\sigma}{|S_t|} \times {\sum_{c \in S_t}\sum_{i \in 1 \ldots m} \mathcal{E}_{t,c}^i} \Big\}
		\label{eq:A_t}
	\end{equation}
	where $\sigma$ is the hyper-tuned parameters for determining the threshold, which varies from task to task.
	
	Finally, each selected client's parameters are assigned an aggregation weight (\( \vartheta_{t,k} \)), reflecting their contribution to the global model update. This weight value is determined by considering the client's accuracy, the number of true positive instances, and the prioritization score of the respective class, among other factors. This is computed as follows:
	\begin{equation}
		\begin{split}
			\vartheta_{t,k} &= \frac{\sum_{i \in 1 \ldots m}\texttt{ACC}_{t,k}[i] \times \sum_{i \in 1 \ldots m}\texttt{TP}_{t,k}[i]  }{\sum_{i \in 1 \ldots m} \mathcal{E}_{t,k}[i] } \\
			&\times \frac{ max(1,\texttt{TP}_{t,k}[\mathcal{PC}_t])}{ \sum_{i \in 1 \ldots m}\texttt{T}_{t,k}[i] - \sum_{i \in 1 \ldots m}\texttt{TP}_{t,k}[i]}
		\end{split}
		\label{eq:vertheat_t}
	\end{equation}
	This approach ensures that clients with higher performance and relevance contribute more significantly to the global model update process, thus improving overall model quality. The overall algorithmic steps are depicted in Algorithm~\ref{alg:weight_computation}.
	
	\begin{algorithm}[t]
		\scriptsize
		\caption{Statistical Weight Computation}
		\label{alg:weight_computation}
		\SetKwInOut{Input}{Input} 
		\SetKwInOut{Output}{Output}
		\SetKwInOut{Procedure}{Procedure}
		\Input{Client parameters learning statistics: $stats$, list of clients who locally trained the model and submitted to the server: $S_t$}
		\Output{Weights: $weights$}
		calculate error and accuracy\\
		\For{each client $k $ in $ S_t$}{
			\For {worker $w$ in $S_{k,t}$}{
				\For{each class $i$ in ($1 \ldots m$)}{
					\textbf{Compute} $\texttt{FNR}_{t,k}$ following Equation~\ref{eq:FNR}.\\
					\textbf{Compute} $\texttt{FPR}_{t,k}$ following  Equation~\ref{eq:FPR}.\\
					\textbf{Compute} $\texttt{ACC}_{t,k}$ following  Equation~\ref{eq:ACC}.\\
					
				}
			}
			
		}
		Analyze the performance.\\
		\For{each client $k $ in $ S_t$}{
			\For{each class $i$ in ($1 \ldots m$)}{
				compute $\mathcal{E}_{t,k}^i$ following Equation~\ref{eq:Ei}.\\
			}
		}
		Choose the priority class $\mathcal{PC}_t$ following the Equation~\ref{eq:PC}.\\
		\textbf{Select} clients parameters set $A_t$ for aggregation following Equation~\ref{eq:A_t}.\\
		\For{each client $k $ in $ A_t$}{
			Compute weight $\vartheta_{t,k}$ following Equation~\ref{eq:vertheat_t}.
		}
	\end{algorithm}
	
	\subsection{Global model update}
	Once the aggregated weights $\vartheta_{t,k}$ is computed, the server applies weighted federated aggregation to update the global model by utilizing global learning rate $\eta_g$, according to the following Equation:
	
	\begin{equation}
		\Theta_{t+1} = \Theta_{t} -  \eta_g \times \sum_{k \in A_t} \vartheta_{t,k} \times (\Theta_{t} - \theta_{t,k}) 
		\label{eq:global_model_update}
	\end{equation}
	
	Observe that, unlike existing aggregation methods, the introduction of prioritized score based weight $\vartheta_{t,k}$ in Equation~\ref{eq:global_model_update} provides more attention to  high-performing clients for priority classes.
	
	\subsection{Training process}
	
	\begin{algorithm}[t]
		\scriptsize
		\caption{\textsf{FedSat} Algorithm}
		\label{algo:proposed}
		\SetKwInOut{Input}{Input} 
		\SetKwInOut{Output}{Output}
		\SetKwInOut{Procedure}{Procedure}
		\Input{Initial global model parameters: $\Theta_0$, Initial global control variate: $c_0$, Set of participated client: $\mathcal{K}$, Hyper-parameters: Global learning rate $\eta_g$, Local learning rate $\eta_l$, number of local epochs  $\mathbf{E}$, Number of global rounds $\mathbf{R}$}
		\Output{Trained model: $\Theta$}
		\For{t in range($\mathbf{R}$)}{
			\textbf{Server selects} a subset $S_t$ of $n$ clients at random.\\
			\textbf{Set} $W_t \leftarrow \mathcal{K} - S_t$\\ 
			\textbf{At client side:}\\
			\For{each client $k$ in $S_t$}{
				\textbf{Fetch} global model parameters $\Theta_t $ and the hyper-parameters ($\eta_l$, $\mathbf{E}$) from the server.\\
				\textbf{Compute} global correction term $v_{t}$ following Equation~\ref{eq:global_correction_term}.\\
				\textbf{Initialize} local model parameters $\theta_{t,k} \leftarrow \Theta_t$\\
				\For{epoch $i$ in range($\mathbf{E}$)}{
					\textbf{Update} the local model parameters following Equation~\ref{eq:local_per_epoch_update}.\\
				}
				\textbf{Send} the locally trained model parameters $\theta_{t,k}$ to the server.
			}
			\For{each client $k$ in $S_t$}{
				\textbf{Server selects} a subset $W_{t,k}$ of $m$ workers at random from set $W_t$ along with client $k$\\
				\textbf{Update} $W_t \leftarrow W_t - W_{t,k} $\\
				\textbf{At worker side:}\\
				\For{each worker $w$ in $W_{t,k}$}{
					\textbf{Fetch} updated parameters $\theta_{t,k}$ of client $k$.\\
					\textbf{Compute} prediction matrix $\mathbf{\hat{Z}^{w}_{t,k}}$ following Equation~\ref{eq:prediction_mx}.\\
					\textbf{Compute} true positive matrix $\mathbf{TP^{w}_{t,k}}$ following Equation~\ref{eq:TP_mx}.\\
					\textbf{Compute} target matrix $\mathbf{T_{t,k}}$ following Equation~\ref{eq:target_mx}.\\
					\textbf{Send} the analysis results to the server.\\
				}
			}
			\textbf{Compute} statistical weight for each client following Algorithm~\ref{alg:weight_computation}.\\
			\textbf{Compute} global model parameters $\Theta_{t+1}$ following Equation~\ref{eq:global_model_update}.\\
		}
		
	\end{algorithm}
	
	The core training process of \textsf{FedSat}, as outlined in Algorithm~\ref{algo:proposed}, is a multi-step iterative approach. At each global round $t$, the server interacts with a subset of clients and their associated workers to update the global model. Let us describe this in detail:
	\begin{enumerate}
		\item \textbf{Initialization.} At the beginning of the training process, the server defines the set of clients $\mathcal{K}$ who participate in the training and statistical analysis process, and sets the following: (a) initial global model parameters \( \Theta_0 \), (b) initial global control variate \( c_0 \), (c) set of participated clients $\mathcal{K}$, and (d) various hyper-parameters. 
		
		\item \textbf{Iterative training.} In each round \( t \), the server randomly selects a subset \( S_t \) of clients to participate in the local training. Each selected client \( k  \in S_t \) fetches the current global model parameters \( \Theta_t \) and computes global correction term $v_t$. It then initializes its local model parameters \( \theta_{t,k} \) with $\Theta_t$ and performs local training for a predetermined number of epochs $\mathbf{E}$. During each epoch $e$, the client updates its model parameters based on its local dataset \( \mathcal{D}_k \), according to Equation~\ref{eq:local_per_epoch_update}. After completing the local training, each client submits its updated model parameters back to the server. Next, for each client $k \in S_t$, the server forms a worker-set \( W_{t,k} \) to perform parameters evaluation. Each worker in set $W_{t,k}$ fetches their associated model parameters from the server and computes learning statistics by evaluating their performance, which are eventually send back to the server. Based on the statistical performance analysis of all the model parameters, the server selects a set \( A_t \) of locally trained parameters and computes aggregation weights $\vartheta_{t,k}$ according to Algorithm \ref{alg:weight_computation}. Finally, the global model parameters (\( \Theta_t \)) are updated using the computed weights.
		
		\item  \textbf{Termination.} After completing \( T \) rounds of training, the training process terminates and the optimal global model $\Theta^*$ is obtained.\\
		\[\Theta^* = \underset{t \in 1 \ldots T}{max} \,  accuracy(\Theta_t)\]
	\end{enumerate}

	\section{Convergence Proof}
	
	\label{sec:convergence proof}
	\begin{assumption}
		\label{assump:1}
		Both the prediction-sensitive loss function $\mathcal{L}_{PS}$ and the cross-entropy loss function $\mathcal{L}_{CE}$ are $\mu$-strongly convex and $L$-smooth.
	\end{assumption}
	
	\begin{assumption}
		\label{assump:2}
		Let $\hat{l}_i$ be the predicted class of an input sample labelled with class $l_i$. The cost incurred during loss minimization by the loss function $\mathcal{L}_{PS}$ is as follows:
		\begin{equation}
			cs_{l_i,\hat{l}_i} =
			\begin{cases}
				1 & \text{if } \hat{l}_i = l_i \\
				1 < cs_{l_i,\hat{l}_i} < C_{max} & \text{if } \hat{l}_i \neq l_i
			\end{cases}
		\end{equation}
		where $C_{max}$ is the maximum cost bound.
	\end{assumption}

	\begin{lemma}[Improvement Through Prediction-Sensitive Loss]
		\label{lemma:1}
		Let $\theta_{PS}$ and $\theta_{CE}$ be the parameters obtained after training with prediction-sensitive loss function $\mathcal{L}_{PS}$ and cross-entropy loss functions $\mathcal{L}_{CE}$, respectively. Let both the assumptions~\ref{assump:1} and~\ref{assump:2} hold. Then, the expected improvement of the local parameters updated using the $\mathcal{L}_{PS}$ compared to the parameters obtained by employing $\mathcal{L}_{CE}$ is always greater than or equal to zero. That is, $\mathbb{E}[\theta_{PS} - \theta_{CE}] \ge 0$.
	\end{lemma}
	
	\begin{prooft}
		Please refer to Proof 1 in Appendix A, available with this manuscript.
	\end{prooft}
	
	\begin{lemma}[Improved Accuracy for Critical Class through Prioritized Class-based Weighted Aggregation Scheme]
		\label{lemma:2}
		Let $Acc_{\mathcal{PC}_t}(\Theta_t)$ be the accuracy of the model $\Theta_t$ on the priority class $\mathcal{PC}_t$ at round $t$. Our proposed prioritized-class based weighted aggregation scheme ensures an expected improvement in accuracy for $\mathcal{PC}_t$ by an amount $\delta$ (where $\delta > 0$), as follows:
		\[
		\mathbb{E}[Acc_{\mathcal{PC}_t}(\Theta_{t+1})] \geq Acc_{\mathcal{PC}_t}(\Theta_t) + \delta
		\]
		
	\end{lemma}
	
	\begin{prooft}
		Please refer to Proof 2 in Appendix A, available with this manuscript.
	\end{prooft}
	
	\begin{lemma}[Convergence of Global Model]
		\label{lemma:3}
		Let $\Theta_t$ be the global model parameters at round $t$. Under Assumptions \ref{assump:1} and \ref{assump:2}, and with appropriate learning rates $\eta_l$ and $\eta_g$, \textsf{FedSat} converges in expectation $\mathbb{E}[\mathcal{L}(\Theta_{t+1})] \leq \mathcal{L}(\Theta_t) - \rho^2$, where $\mathcal{L}(\Theta)$ is the global model loss function and $\rho$ is a constant representing the improvement per round.
	\end{lemma}
	
	\begin{prooft}
		Please refer to Proof 3 in Appendix A, available with this manuscript.
	\end{prooft}
	
	\begin{theorem}[Faster Convergence and Robust Model Generation in \textnormal{\textsf{FedSat}}]
		Let $\{\Theta_t\}_{t=0}^T$ be the sequence of global model parameters produced by \textsf{FedSat} over $T$ rounds using the prediction-sensitive loss function $\mathcal{L}_{PS}$. Let $\{\Theta'_t\}_{t=0}^T$ be the sequence produced by a standard federated learning algorithm under the same conditions using the standard cross-entropy loss function $\mathcal{L}_{CE}$. Under assumptions \ref{assump:1} and \ref{assump:2}, \textsf{FedSat} guarantees faster convergence and robust global model generation over sufficiently large $T$, formalized as follows:
		
		\emph{Faster convergence:} $\exists \gamma > 1$ such that
		\[
		\mathbb{E}[\mathcal{L}_{PS}(\Theta_0) - \mathcal{L}_{PS}(\Theta_T)] \geq \gamma \cdot \mathbb{E}[\mathcal{L}_{CE}(\Theta'_0) - \mathcal{L}_{CE}(\Theta'_T)]
		\]
		
		\emph{Robust model generation:} $\exists \varrho > 0$ such that $\forall t \in \{1,\ldots,T\}$: $\mathbb{E}[Acc_{\mathcal{PC}_t}(\Theta_t)] \geq \mathbb{E}[Acc_{\mathcal{PC}_t}(\Theta'_t)] + \varrho$
	\end{theorem}
	
	\begin{prooft}
		This is proved based on three key Lemmas~\ref{lemma:1}, \ref{lemma:2}, and \ref{lemma:3}. Please refer to the Proof 4 in Appendix A for complete proof, available with this manuscript.
	\end{prooft}
	
	\section{Experimental Evaluation}
	\label{sec:implementation and evaluation}
	
	This section presents performance evaluation of \textsf{FedSat}, with a detailed comparative analysis w.r.t. the baseline models. 
	
	\subsection{Implementation}
	We implement \textsf{FedSat} and the baseline methods using Python 3.9 with the support of PyTorch library. 
	For local training, we consider four neural networks, namely MLP, LeNet-5, ResNet-8, and ResNet-18. Their architectures are described as follows: (1) MLP model is a 3-layer network with neurons (80, 60, 10) and ELU activation; (2) LeNet-5~\cite{lenet5} model is a 7-layer CNN with 5x5 convolutions, tanh activation, and average pooling; (3) ResNet-8~\cite{he2016deep} model is a CNN with an initial 3x3 convolutional layer, followed by 3 residual blocks, batch normalization, ReLU activation, and global average pooling; and (4) ResNet-18~\cite{he2016deep} model is a CNN with an initial 7x7 convolutional layer, followed by 4 residual blocks, batch normalization, ReLU activation, and global average pooling. We employed the PyTorch SGD optimizer for updating model parameters during training.
	
	\subsection{Experiment Setup}
	In this study, we conduct a comprehensive evaluation of \textsf{FedSat}, by comparing its performance against state-of-the-art approaches. To this aim, we consider the following nine baseline methods: FedAvg, FedProx, MOON, SCAFFOLD, FedLC, FedAvgM, FedLada, Elastic, and FedDyn, which represent a diverse range of strategies for federated learning. Our experiments are performed on three widely-used datasets: MNIST, CIFAR-10, and CIFAR-100, each presenting unique challenges and characteristics. The details of these datasets are summarized in Table \ref{tab:datasets}.
	To investigate the effectiveness of \textsf{FedSat} and the baseline methods under different model architectures, we employ a variety of neural networks for local training. Specifically, when training on the MNIST dataset, we utilize MLP and LeNet-5 models, which are well-suited for the task of handwritten digit recognition. For more complex CIFAR-10 and CIFAR-100 datasets, we employ ResNet-8 and ResNet-18 models, which already depicted a superior performance on image classification tasks. To ensure fairness and reproducibility of our comparisons, we maintain fixed random seeds and consistent settings across all experiments. This allows us to isolate the impact of federated learning algorithms on model performance, minimizing the influence of random variations. 
	
	\subsection{Dataset Distribution Settings}
	To explore the performance under various data distribution scenarios, we establish the following three distinct client dataset configurations:
	\begin{enumerate}
		\item \emph{Label skewed} (\textbf{LS}): In this setting, the label distribution varies across clients, simulating a scenario where each client has a different proportion of samples from each class. To create a label-skewed dataset, we use the Dirichlet distribution \cite{Dirichlet} on the label ratios to ensure uneven label distributions among clients.
		\item \emph{Label skew with missing classes} (\textbf{LSMC}): In this setting, we create a non-IID dataset where, in addition to the label skewness, each client has a different subset of the total classes, and the distributions of these classes are skewed. To achieve this, we randomly select $C$ classes out of $m$ available classes, where $C < m$. Lower values of $C$ indicate a more extreme non-IID setting. This configuration simulates a more challenging scenario where some classes are entirely absent from clients' data distributions. This reflects real-world situations where clients may have access to only a subset of the overall class distribution, and the available classes are not evenly represented.
		\item \emph{Label and Quantity Skew with Missing Classes} (\textbf{LQSMC}): 
		In this case, along with the previous data setting (\textbf{LSMC}), we additionally introduce quantity skewness in the data distribution across the clients. Specifically, we determine the number of classes available for each client by randomly sampling from the range [$n$, $m$], where $n$ is the minimum number of classes per client and $m$ is the total number of classes in the dataset. To incorporate quantity skew, we assign a varying number of samples to each client based on their available number of classes, with class ratios skewed within the available classes. Clients with a higher number of classes receive a larger quantity of samples, while clients with fewer classes receive a smaller quantity.
		
	\end{enumerate}
	By considering these diverse dataset configurations, we aim to provide a thorough and realistic assessment of \textsf{FedSat} and the baseline methods, revealing their strengths and limitations under a range of non-IID data distributions. 
	
	\begin{table}[t]
		\centering
		\caption{Details of Datasets}
		\label{tab:datasets}
		\scriptsize
		\resizebox{\columnwidth}{!}{
			\begin{tabular}{|c|c|c|c|c|}
				\hline
				\textbf{Name}   & \textbf{Train Samples} & \textbf{Test Samples} & \textbf{Input Size} & \textbf{Classes} \\ \hline
				MNIST           & 60,000                 & 10,000                & 28 $\times$ 28               & 10               \\ \hline
				CIFAR-10        & 50,000                 & 10,000                & 32 $\times$ 32               & 10               \\ \hline
				CIFAR-100       & 50,000                 & 10,000                & 32 $\times$ 32               & 100              \\ \hline
			\end{tabular}
		}
	\end{table}
	
	\begin{table*}[htbp]
		\centering
		\caption{Comparison of method performance for training with 16 batch size ($\mathbf{B}$) and 5 local epochs ($\mathbf{E}$) on different datasets and model architecture in label-skewed data settings. Here, $\eta_l$ represents the local learning rate used for training.}
		\label{tab:dirichlet_results}
		\resizebox{\textwidth}{!}{%
			\begin{tabular}{@{}l|cc|cc|cc|cc|cc|cc@{}}
				\toprule
				\multirow{4}{*}{\textbf{Methods}} & \multicolumn{6}{c|}{\textbf{$\eta_l$=0.01}}  & \multicolumn{6}{c}{\textbf{$\eta_l$=0.001}} \\ 
				\cmidrule(r){2-7} \cmidrule(l){8-13}
				& \multicolumn{2}{c|}{MNIST} & \multicolumn{2}{c|}{CIFAR-10} & \multicolumn{2}{c|}{CIFAR-100} & \multicolumn{2}{c|}{MNIST} & \multicolumn{2}{c|}{CIFAR-10} & \multicolumn{2}{c}{CIFAR-100} \\ 
				\cmidrule(r){2-3} \cmidrule(lr){4-5} \cmidrule(lr){6-7} \cmidrule(lr){8-9} \cmidrule(lr){10-11} \cmidrule(l){12-13}
				& MLP & LeNet-5 & ResNet-8 & ResNet-18 &ResNet-8 & ResNet-18 &  MLP & LeNet-5 & ResNet-8 & ResNet-18 &ResNet-8 & ResNet-18 \\ 
				\midrule
				Elastic &95.62&98.76&76.82&83.95&38.34&57.53&90.89&98.45&63.96&72.60&21.15&31.49\\
				FedAvg &95.14&98.48&75.87&84.02&36.67&57.33&90.56&98.12&62.14&71.50&18.68&30.28\\
				FedAvgM &96.08&98.59&74.71&85.20&35.37&58.61&91.79&97.88&59.18&78.07&20.81&37.74\\
				FedDyn &92.15&96.72&64.78&82.58&20.78&55.62&90.71&97.56&56.30&77.11&20.96&36.70\\
				FedLADA &95.72&98.85&75.02&79.11&31.17&44.22&95.96&97.23&59.90&75.14&25.44&27.61\\
				FedLC &95.00&98.41&70.76&83.78&27.87&57.22&90.44&97.76&56.16&70.94&16.43&27.66\\
				FedProx &95.28&98.27&61.52&83.91&25.94&51.34&90.57&97.55&62.26&71.74&18.55&30.32\\
				MOON &96.28&98.57&77.12&86.31&39.92&58.46&94.41&98.18&62.56&74.74&22.79&33.54\\
				SCAFFOLD &98.28&99.09&86.07&88.61&58.62&66.21&96.33&98.33&82.58&84.41&43.91&55.12\\
				\textsf{FedSat} &\textbf{98.51}&\textbf{99.47}&\textbf{87.67}&\textbf{90.70}&\textbf{60.35}&\textbf{68.48}&\textbf{97.00}&\textbf{98.67}&\textbf{84.05}&\textbf{85.61}&\textbf{52.35}&\textbf{58.79}\\
				\bottomrule
			\end{tabular}%
		}
	\end{table*}
	
	\subsubsection{Evaluation on \textnormal{\textbf{LS}}}
	
	The performance evaluation of \textsf{FedSat} w.r.t. the baselines on \textbf{LS} distribution is summarized in Table~\ref{tab:dirichlet_results}. As observed, \textsf{FedSat} consistently demonstrates higher test accuracy than the baseline methods across all different datasets and model architectures. On an average, \textsf{FedSat} surpasses the second-best method, SCAFFOLD, by 1.8\% and the weakest-performing method, FedLC, by 19.87\%. These results underscore \textsf{FedSat}'s effectiveness in addressing label-skewed non-IID settings, potentially due to its novel prediction-sensitive loss function and proposed weighted aggregation mechanism which incorporates both parameters quality and imbalance training. Further, the sensitivity to local learning rates, as observed in case of FedProx, FedDyn, and FedLADA, emphasizes the critical need for meticulous hyper-parameter tuning.  
	
	\begin{table*}[htbp]
		\centering
		\caption{Comparison of method performance for training with ($\eta_l$=0.01, $\mathbf{B}$=16, $\mathbf{E}$=5) on label skewed datasets with missing class data settings. Here, $n$ denotes the number of classes present in each client.}
		\label{tab:label_skew_results}
		\resizebox{\textwidth}{!}{%
			\begin{tabular}{@{}l|ccc|ccc|ccc|ccc|ccc|ccc@{}}
				\toprule
				\multirow{4}{*}{\textbf{Methods}} & \multicolumn{6}{c|}{MNIST} & \multicolumn{6}{c|}{CIFAR-10} & \multicolumn{6}{c}{CIFAR-100}  \\ 
				\cmidrule(r){2-7} \cmidrule(lr){8-13} \cmidrule(l){14-19}
				&  \multicolumn{3}{c|}{MLP} & \multicolumn{3}{c|}{LeNet-5} & \multicolumn{3}{c|}{ResNet-8} & \multicolumn{3}{c|}{ResNet-18} & \multicolumn{3}{c|}{ResNet-8} & \multicolumn{3}{c}{ResNet-18} \\ 
				\cmidrule(r){2-4} \cmidrule(lr){5-7} \cmidrule(lr){8-10} \cmidrule(r){11-13} \cmidrule(lr){14-16} \cmidrule(l){17-19}
				& $n$=2 & $n$=4 & $n$=6 & $n$=2 & $n$=4 & $n$=6 & $n$=2 & $n$=4 & $n$=6 & $n$=2 & $n$=4 & $n$=6 & $n$=20 & $n$=40 & $n$=60 & $n$=20 & $n$=40 & $n$=60 \\
				\midrule
				Elastic  & 90.43 & 93.22 & 94.49 & 96.32 & 97.99 & 98.06 & 57.46 & 63.52 & 66.09 & 45.1 & 64.52 & 72.87 & 33.37 & 38.23 & 38.31 & 57.81 & 61.03 & 60.25 \\
				FedAvg   & 90.1  & 92.84 & 93.92 & 95.92 & 97.74 & 97.76 & 55.66 & 62.03 & 65.49 & 43.04 & 62.6 & 71.66 & 32.04 & 37.01 & 37.53 & 55.99 & 60.81 & 59.55 \\
				FedAvgM  & 92.61 & 93.62 & 94.82 & 96.69 & 97.78 & 98.15 & 45.29 & 58.75 & 63.3 & 42.53 & 53.57 & 63.13 & 22.43 & 29.22 & 32.77 & 38.25 & 50.34 & 58.3 \\
				FedDyn   & 84.72 & 86.7 & 87.55 & 90.37  & 91.15 & 93.08 & 54.55 & 53.73 & 55.8 & 37.24 & 55.25 & 68.68 & 24.31 & 28.42 & 29.58 & 55.56 & 58.55 & 58.19 \\
				FedLADA  & 92.95 & 93.58 & 94.64 & 85.18 & 97.58 & 98.48 & 26.88 & 37.34 & 58.16 & 27.79 & 28.66 & 44.09 & 23.39 & 24.26 & 31.43 & 41.51 & 45.06 & 47.02 \\
				FedLC    & 88.65 & 92.68 & 93.18 & 94.2 & 96.91 & 97.23 & 41.55 & 58.42 & 65.27 & 27.84 & 52.05 & 65.29 & 24.47 & 27.47 & 27.94 & 45.47 & 54.57 & 54.26 \\
				FedProx  & 90.44 & 93.07 & 93.16 & 94.07 & 97.25 & 97.55  & 42.14 & 47.56 & 53.63 & 21.98 & 52.38 & 61.98 & 20.84 & 25.15 & 25.52 & 42.13 & 45.86 & 46.78 \\
				MOON  & 90.89 & 93.37 & 94.36 & 96.58 & 97.93 & 98.09  & 44.53 & 65.16 & 65.19 & 45.96 & 64.67 & 72.84 & 33.61 & 38.14 & 37.52 & 58.13 & 60.86 & 62.78 \\
				SCAFFOLD & 97.91 & 98.00 & 98.45 & 99.21 & 99.31 & 99.46 & 78.33 & 81.25 & 82.29 & 82.81 & 86.95 & 85.68 & 51.28 & 58.51 & 58.66 & 53.7 & 61.83 & 69.24 \\
				\textsf{FedSat} & \textbf{98.49} & \textbf{98.27} & \textbf{98.52} & \textbf{99.44} & \textbf{99.46} & \textbf{99.58} & \textbf{82.87} & \textbf{84.1} & \textbf{84.79} & \textbf{86.31} & \textbf{87.82} & \textbf{85.85} & \textbf{56.13} & \textbf{62.12} & \textbf{62.34} & \textbf{70.27} & \textbf{71.19} & \textbf{70.35} \\
				
				\bottomrule
			\end{tabular}
		}
	\end{table*}
	
	\subsection{Hyper-parameters}
	To ensure a fair and comprehensive evaluation, we carefully select and fix the hyper-parameters for all methods in our experiments. The global learning rate $\eta_g$ is set to 1.0 across all settings, providing a consistent baseline for comparison. For \textsf{FedSat}, 
	we set $\epsilon_1 = 1$ and $\epsilon_2 = 2$ to control the loss refinement mechanism of prediction-sensitive loss function. To calculate the prioritization score of each client's parameters, we determine $\alpha=0.3$ and $\beta=0.2$, giving appropriate weight to both the client's data characteristics and model performance.
	
	For the baseline methods, we consider optimal hyper-parameters settings as recommended in their respective proposals. To maintain consistency and fairness across all methods, we fix the local learning rate $\eta_l$, batch size $\mathbf{B}$, number of epochs $\mathbf{E}$, and number of global rounds $\mathbf{R}$. We conduct experiments with two local learning rates, $\eta_l \in \{0.01, 0.001\}$, and two batch size settings, $\mathbf{B} \in \{16, 32\}$, to explore the impact of these hyper-parameters on the performance of each method. By carefully controlling these variables, we aim to provide a rigorous and unbiased evaluation of \textsf{FedSat} and the baseline approaches, shedding light on their relative strengths and weaknesses under various hyper-parameter configurations.
	
	\subsection{Performance Evaluation}
	In this section, we demonstrate the performance of \textsf{FedSat} under various heterogeneous settings compared to the state-of-the-art solutions. We perform 200 rounds for  MLP and LeNet-5 MNIST, 400 rounds for ResNet-8 on CIFAR-10 and CIFAR-100, and 250 rounds for ResNet-18 on CIFAR-10 and CIFAR-100 datasets. The results under three different heterogeneous data-distributions are described below:
	
	\subsubsection{Evaluation on \textnormal{\textbf{LSMC}}}
	The performance in case of label skewed datasets with missing classes is depicted in Table~\ref{tab:label_skew_results}, where the lower value of $n$ indicates more challenging heterogeneous settings. 
	
	Observed that \textsf{FedSat} achieves highest accuracy in all settings, with an average improvement of 3.34\% over the second-best method. It is worthwhile to mention that \textsf{FedSat} demonstrates more significant improvements on challenging datasets, achieving a 0.51\% improvement on the simplest dataset MNIST, a 1.98\% improvement on CIFAR-10, and a remarkable 6.86\% improvement on the most challenging dataset CIFAR-100 compared to the best baseline method. Furthermore, in the extreme heterogeneous setting ($n$=2), \textsf{FedSat} exhibits a more pronounced impact on accuracy, with an average improvement of 4.51\% across all settings compared to the best-performing baseline SCAFFOLD.
	
	This suggests that \textsf{FedSat}'s design elements, particularly the prediction-sensitive loss function, effectively mitigate the negative effects of label skewed data distribution and our novel prioritized-class based weighted aggregation scheme dealing with the unstable trained model due to missing classes. In contrast, the methods based on simple aggregation schemes, such as FedAvg, FedAvgM, FedLC, and FedProx, experience substantial performance drops, emphasizing the necessity for adaptations to handle the impact of missing classes. Interestingly, when training with a high local learning rate (0.01), FedProx, FedDyn, and FedLADA exhibit significant performance drops and instability mainly due to their additional gradient correction being proportional to the local learning rate.

	\begin{table*}[t]
		\centering
		\caption{Comparison of method performance for training with ($\eta_l$=0.01, $\mathbf{B}$=16, $\mathbf{E}$=5) on combination of label skewed and quantity skewed data settings. Here, $n$ denotes the minimum number of classes present in each client.}
		\tiny
		\resizebox{\textwidth}{!}{%
			\begin{tabular}{@{}l|cc|cc|cc|cc|cc|cc@{}}
				\toprule
				\multirow{4}{*}{\textbf{Methods}} & \multicolumn{4}{c|}{MNIST} & \multicolumn{4}{c|}{CIFAR-10} & \multicolumn{4}{c}{CIFAR-100}  \\ 
				\cmidrule(r){2-5} \cmidrule(lr){6-9} \cmidrule(l){10-13}
				&  \multicolumn{2}{c|}{MLP} & \multicolumn{2}{c|}{LeNet-5} & \multicolumn{2}{c|}{ResNet-8} & \multicolumn{2}{c|}{ResNet-18} & \multicolumn{2}{c|}{ResNet-8} & \multicolumn{2}{c}{ResNet-18} \\ 
				\cmidrule(r){2-5} \cmidrule(lr){6-9} \cmidrule(l){10-13}
				& $n$=2 & $n$=4 & $n$=2 & $n$=4 & $n$=2 & $n$=4 & $n$=2 & $n$=4 & $n$=20 & $n$=40 & $n$=20 & $n$=40  \\
				\midrule
				Elastic  & 94.93 & 98.53 & 98.53 & 98.31 & 66.64 & 69.42 & 72.25 & 73.51 & 35.87 & 37.27 & 55.95 & 54.33 \\
				FedAvg   & 94.29 & 98.24 & 98.24 & 97.97 & 65.66 & 68.13 & 70.92 & 73.68 & 34.35 & 36.16 & 55.84 & 53.88 \\
				FedAvgM  & 94.62 & 98.66 & 98.66 & 98.03 & 62.51 & 63.14 & 65.34 & 65.89 & 27.89 & 31.37 & 58.4  & 58.93 \\
				FedDyn   & 87.26 & 90.32 & 92.73 & 93.08 & 54.65 & 56.43 & 67.52 & 68.65 & 21.27 & 22.18 & 54.19 & 53.43 \\
				FedLADA  & 93.45 & 98.96 & 98.96 & 98.22 & 54.74 & 59.76 & 44.86 & 53.74 & 29.03 & 30.23 & 24.73 & 37.74 \\
				FedLC    & 93.78 & 98.05 & 98.05 & 97.37 & 56.94 & 59.38 & 65.38 & 67.54 & 26.25 & 26.37 & 51.61 & 50.66 \\
				FedProx  & 94.48 & 98.05 & 98.15 & 97.37 & 53.55 & 54.50  & 61.43 & 65.76 & 23.63 & 24.43 & 52.98   & 41.47   \\
				MOON  & 94.84 & 98.35 & 98.25 & 98.06 & 67.55 & 68.54  & 71.79 & 73.76 & 36.63 & 38.62 & 56.35   & 55.89   \\
				SCAFFOLD & 98.24 & 98.21 & 99.12 & \textbf{99.57} & 82.91 & 84.38 & \textbf{85.74} & 86.35 & 56.13 & 56.57 & 64.39 & 63.13  \\
				\textsf{FedSat}     & \textbf{98.54} & \textbf{98.58} & \textbf{99.46} & 99.52 & \textbf{84.08} & \textbf{85.07} & 85.68 & \textbf{86.89} & \textbf{58.85} & \textbf{60.21} & \textbf{65.64} & \textbf{64.96} \\
				
				\bottomrule
			\end{tabular}
		}
		\label{tab:mixed_skew_results}
	\end{table*}
	
	\subsubsection{Evaluation on \textnormal{\textbf{LQSMC}}}
	
	Table~\ref{tab:mixed_skew_results} illustrates the performance comparison of \textsf{FedSat} against baselines under the combined challenges of label skew and quantity skew data distribution with missing classes. The results demonstrate that \textsf{FedSat} achieves the highest accuracy in most settings, especially on the more challenging CIFAR-10 and CIFAR-100 datasets. On an average, \textsf{FedSat} outperforms the second-best method by 1.47\% across all settings. It is worth to mention that, methods based on simple aggregation schemes, like FedAvg, FedAvgM, FedProx and FedLC, experience substantial performance drops, with an average decrease of 19.81\% compared to \textsf{FedSat}.
	Notably, as the dataset complexity increases, \textsf{FedSat}'s performance improvement becomes more significant. On MNIST, CIFAR-10, and CIFAR-100, FedSat outperforms the best alternative, SCAFFOLD, by average margins of 0.23\%, 1.18\%, and 2.03\%, respectively.
	Similar to \textbf{LSMC}, the performance drops of  FedProx, FedDyn, and FedLADA are also observed, due to training with high local learning rate. 
	
	This significant advantage suggests that \textsf{FedSat}'s design choices, particularly the proposed weighted aggregation mechanism that considers both data quality and quantity, are highly effective in mitigating the negative effects of model biases towards a class due to data imbalances.
	
	\subsection{\textsf{FedSat's} Dependability Analysis}
	
	\begin{figure*}[t]
		\centering
		\begin{tabular}{ccc}
			
			\scalebox{0.52}{
				\begin{tikzpicture}
					\begin{axis}[
						width=10cm,
						height=8cm,
						ybar,
						bar width=0.30cm,
						legend style={
							at={(0.5,-0.15)},
							anchor=north,
							legend columns=5,
							font=\small,
							/tikz/every even column/.append style={column sep=0.5cm}
						},
						enlarge x limits=0.55,
						ylabel={Accuracy (\%)},
						symbolic x coords={Min-Acc, Max-Acc},
						xtick=data,
						nodes near coords,
						every node near coord/.append style={font=\small, rotate=90, anchor=west, /pgf/number format/.cd, fixed zerofill, precision=2},
						ymin=0, ymax=1.2,
						ymajorgrids=true,
						grid style=dashed,
						tick label style={font=\small},
						label style={font=\small},
						]
						
						\addplot[fill=olive] coordinates {(Min-Acc, 0.01) (Max-Acc, 0.85)};
						\addplot[fill=blue] coordinates {(Min-Acc, 0.01) (Max-Acc, 0.92)};
						\addplot[fill=cyan] coordinates {(Min-Acc, 0.11) (Max-Acc, 0.85)};
						\addplot[fill=orange] coordinates {(Min-Acc, 0.01) (Max-Acc, 0.88)};
						\addplot[fill=green] coordinates {(Min-Acc, 0.01) (Max-Acc, 0.88)};
						\addplot[fill=red] coordinates {(Min-Acc, 0.04) (Max-Acc, 0.62)};
						\addplot[fill=purple] coordinates {(Min-Acc, 0.01) (Max-Acc, 0.92)};
						\addplot[fill=magenta] coordinates {(Min-Acc, 0.01) (Max-Acc, 0.93)};
						\addplot[fill=brown] coordinates {(Min-Acc, 0.36) (Max-Acc, 0.96)};
						\addplot[fill=black    ] coordinates {(Min-Acc, 0.41) (Max-Acc, 1.0)};
						
						\legend{
							Elastic,
							FedAvg, 
							FedLC,
							FedAvgM, 
							FedDyn, 
							FedLADA, 
							FedProx,
							MOON, 
							SCAFFOLD,
							\textsf{FedSat},
						}
					\end{axis}
				\end{tikzpicture}
			} &
			
			\scalebox{0.52}{
				\begin{tikzpicture}
					\begin{axis}[
						width=10cm,
						height=8cm,
						ybar,
						bar width=0.30cm,
						legend style={
							at={(0.5,-0.15)},
							anchor=north,
							legend columns=5,
							font=\small,
							/tikz/every even column/.append style={column sep=0.5cm}
						},
						enlarge x limits=0.55,
						ylabel={Accuracy (\%)},
						symbolic x coords={Min-Acc, Max-Acc},
						xtick=data,
						nodes near coords,
						every node near coord/.append style={font=\small, rotate=90, anchor=west, /pgf/number format/.cd, fixed zerofill, precision=2},
						ymin=0, ymax=1.2,
						ymajorgrids=true,
						grid style=dashed,
						tick label style={font=\small},
						label style={font=\small},
						]
						
						\addplot[fill=olive] coordinates {(Min-Acc, 0.01) (Max-Acc, 0.85)};
						\addplot[fill=blue] coordinates {(Min-Acc, 0.01) (Max-Acc, 0.92)};
						\addplot[fill=cyan] coordinates {(Min-Acc, 0.12) (Max-Acc, 0.82)};
						\addplot[fill=orange] coordinates {(Min-Acc, 0.01) (Max-Acc, 0.88)};
						\addplot[fill=green] coordinates {(Min-Acc, 0.01) (Max-Acc, 0.88)};
						\addplot[fill=red] coordinates {(Min-Acc, 0.01) (Max-Acc, 0.92)};
						\addplot[fill=purple] coordinates {(Min-Acc, 0.01) (Max-Acc, 0.92)};
						\addplot[fill=magenta] coordinates {(Min-Acc, 0.01) (Max-Acc, 0.93)};
						\addplot[fill=brown] coordinates {(Min-Acc, 0.16) (Max-Acc, 1.0)};
						\addplot[fill=black    ] coordinates {(Min-Acc, 0.41) (Max-Acc, 1.0)};
						
						\legend{
							Elastic,
							FedAvg, 
							FedLC,
							FedAvgM, 
							FedDyn, 
							FedLADA, 
							FedProx,
							MOON, 
							SCAFFOLD,
							\textsf{FedSat},
						}
					\end{axis}
				\end{tikzpicture}
			} 
			&
			\scalebox{0.52}{
				\begin{tikzpicture}
					\begin{axis}[
						width=10cm,
						height=8cm,
						ybar,
						bar width=0.30cm,
						legend style={
							at={(0.5,-0.15)},
							anchor=north,
							legend columns=5,
							font=\small,
							/tikz/every even column/.append style={column sep=0.5cm}
						},
						enlarge x limits=0.55,
						ylabel={Accuracy (\%)},
						symbolic x coords={Min-Acc, Max-Acc},
						xtick=data,
						nodes near coords,
						every node near coord/.append style={font=\small, rotate=90, anchor=west, /pgf/number format/.cd, fixed zerofill, precision=2},
						ymin=0, ymax=1.2,
						ymajorgrids=true,
						grid style=dashed,
						tick label style={font=\small},
						label style={font=\small},
						]
						
						\addplot[fill=olive] coordinates {(Min-Acc, 0.04) (Max-Acc, 0.96)};
						\addplot[fill=blue] coordinates {(Min-Acc, 0.01) (Max-Acc, 0.96)};
						\addplot[fill=cyan] coordinates {(Min-Acc, 0.16) (Max-Acc, 0.91)};
						\addplot[fill=orange] coordinates {(Min-Acc, 0.01) (Max-Acc, 0.92)};
						\addplot[fill=green] coordinates {(Min-Acc, 0.01) (Max-Acc, 0.96)};
						\addplot[fill=red] coordinates {(Min-Acc, 0.01) (Max-Acc, 1.0)};
						\addplot[fill=purple] coordinates {(Min-Acc, 0.01) (Max-Acc, 1.0)};
						\addplot[fill=magenta] coordinates {(Min-Acc, 0.01) (Max-Acc, 1.0)};
						\addplot[fill=brown] coordinates {(Min-Acc, 0.55) (Max-Acc, 1.0)};
						\addplot[fill=black    ] coordinates {(Min-Acc, 0.70) (Max-Acc, 1.0)};
						
						\legend{
							Elastic,
							FedAvg, 
							FedLC,
							FedAvgM, 
							FedDyn, 
							FedLADA, 
							FedProx, 
							MOON,
							SCAFFOLD,
							\textsf{FedSat},
						}
					\end{axis}
				\end{tikzpicture}
			}  \\
			(a) & (b) & (c)\\
			$\eta_l$-0.001,$\mathbf{B}$-16,$\mathbf{R}$-200,\#C-2 & $\eta_l$-0.001,$\mathbf{B}$-16,$\mathbf{R}$-400,\#C-2 &  $\eta_l$-0.01,$\mathbf{B}$-16,$\mathbf{R}$-400,\#C-2 \\
			$\mathbf{D}$-MNIST,$\mathbf{M}$-LeNet-5 & $\mathbf{D}$-CIFAR-10,$\mathbf{M}$-ResNet-8 & $\mathbf{D}$-CIFAR-100,$\mathbf{M}$-ResNet-18\\
		\end{tabular}
		\caption{Comparison of the highest and lowest client accuracies achieved by \textsf{FedSat} and baseline methods across various settings. The settings are defined by the local learning rate ($\eta_l$), batch size ($\mathbf{B}$), number of global rounds ($\mathbf{R}$), number of classes per client (\#C), dataset ($\mathbf{D}$), and model architecture ($\mathbf{M}$).}
		\label{fig:client-accuracy_comparison}
	\end{figure*}
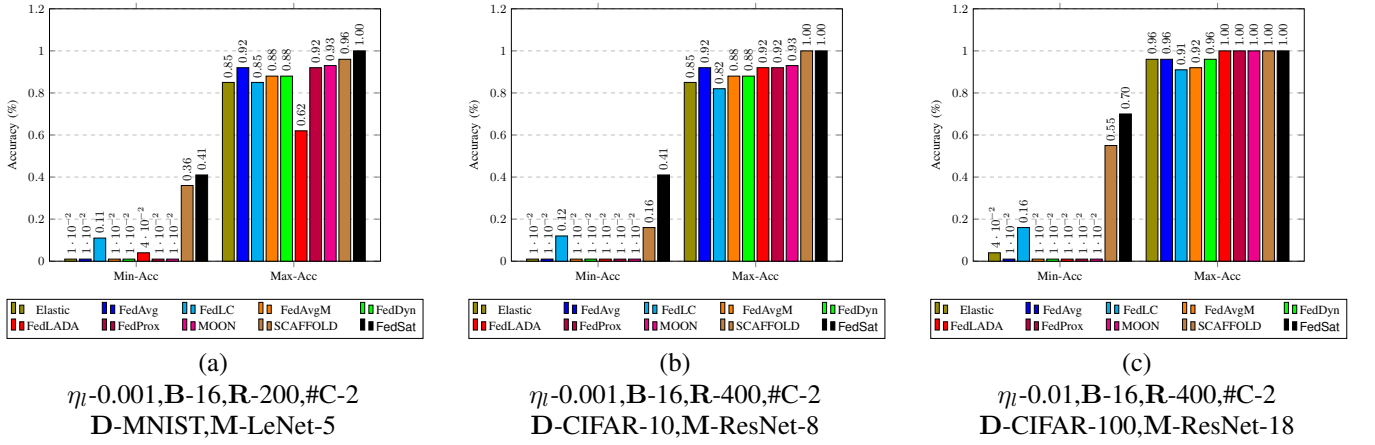
	
	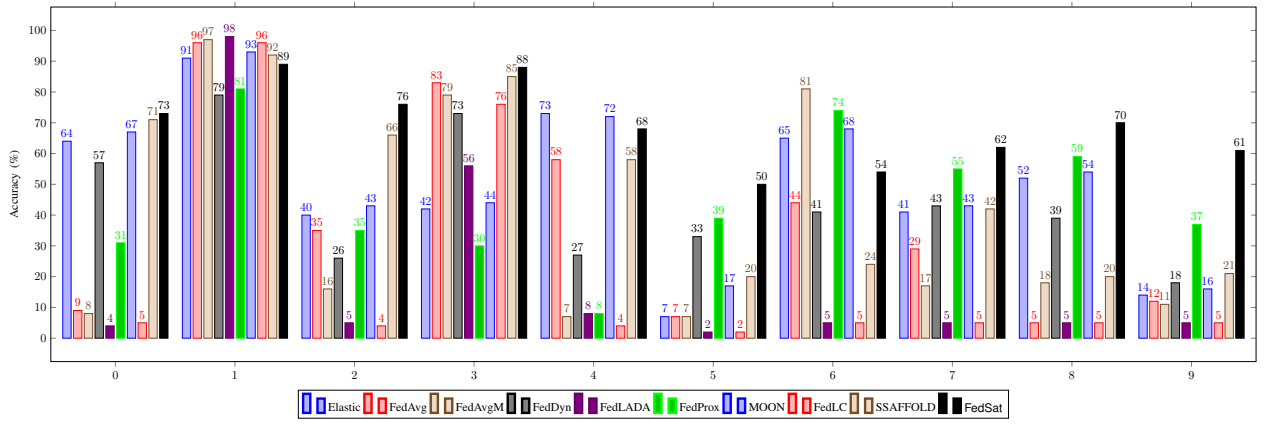
\begin{figure*}[t]
		\centering
		\scalebox{0.45}{
			\begin{tikzpicture}
				\begin{axis}[
					width=37cm,
					height=12cm,
					ybar,
					bar width=0.07,
					legend style={at={(0.5,-0.08)},
						anchor=north,legend columns=-1, legend image post style={scale=2, line width=2pt}},
					enlarge x limits={0.06}, 
				ylabel={Accuracy (\%)},
				xtick=data,
				nodes near coords,
				nodes near coords align={vertical},
				]
				
				\addplot coordinates {(0, 64) (1, 91) (2, 40) (3, 42) (4, 73) (5, 7) (6, 65) (7, 41) (8, 52) (9, 14)};
				
				\addplot coordinates {(0, 9) (1, 96) (2, 35) (3, 83) (4, 58) (5, 7) (6, 44) (7, 29) (8, 5) (9, 12)};
				
				\addplot coordinates {(0, 8) (1, 97) (2, 16) (3, 79) (4, 7) (5, 7) (6, 81) (7, 17) (8, 18) (9, 11)};
				
				\addplot coordinates {(0, 57) (1, 79) (2, 26) (3, 73) (4, 27) (5, 33) (6, 41) (7, 43) (8, 39) (9, 18)};
				
				\addplot coordinates {(0, 4) (1, 98) (2, 5) (3, 56) (4, 8) (5, 2) (6, 5) (7, 5) (8, 5) (9, 5)};
				
				\addplot coordinates {(0, 31) (1, 81) (2, 35) (3, 30) (4, 8) (5, 39) (6, 74) (7, 55) (8, 59) (9, 37)};
				
				\addplot coordinates {(0, 67) (1, 93) (2, 43) (3, 44) (4, 72) (5, 17) (6, 68) (7, 43) (8, 54) (9, 16)};

				\addplot coordinates {(0, 5) (1, 96) (2, 4) (3, 76) (4, 4) (5, 2) (6, 5) (7, 5) (8, 5) (9, 5)};
				
				\addplot coordinates {(0, 71) (1, 92) (2, 66) (3, 85) (4, 58) (5, 20) (6, 24) (7, 42) (8, 20) (9, 21)};
				
				\addplot[fill=black    ] coordinates {(0, 73) (1, 89) (2, 76) (3, 88) (4, 68) (5, 50) (6, 54) (7, 62) (8, 70) (9, 61)};
				
				\legend{Elastic, FedAvg, FedAvgM, FedDyn, FedLADA, FedProx, MOON, FedLC, SSAFFOLD, \textsf{FedSat}}
			\end{axis}
		\end{tikzpicture}
	}
	\caption{Class-wise accuracy comparison between \textsf{FedSat} and baseline methods on the CIFAR-10 dataset with label skew, where each client is assigned only 2 classes.}
	\label{fig:class_wise-accuracy_comparison}
\end{figure*}

\subsubsection{\textbf{Robustness}}
We assess the robustness of our proposed method by comparing the class-wise accuracy with various baselines, as depicted in Figures~\ref{fig:client-accuracy_comparison} and~\ref{fig:class_wise-accuracy_comparison}. It is evident from Figure~\ref{fig:client-accuracy_comparison} that \textsf{FedSat} outperforms all other baselines in terms of minimum testing accuracy. For instance, in Figure~\ref{fig:client-accuracy_comparison}(a), \textsf{FedSat} achieves a minimum accuracy of 0.41, while the highest minimum accuracy among baselines is 0.36. This represents a 13.89\% improvement in accuracy for the lowest-performing clients.
Similarly, in Figure~\ref{fig:client-accuracy_comparison}(b), \textsf{FedSat}'s minimum accuracy is 0.41, compared to 0.16 for the best-performing baseline, indicating a 156.25\% improvement. In the most challenging setting, as observed in Figure~\ref{fig:client-accuracy_comparison}(c), \textsf{FedSat} maintains a minimum accuracy of 0.70, while the highest baseline achieves only 0.55, representing a 27.27\% improvement. 

On the other hand, Figure~\ref{fig:class_wise-accuracy_comparison} demonstrates that, in contrast to the baselines, \textsf{FedSat} achieves consistent accuracy across all classes by leveraging its novel prioritized-class based weighted aggregation scheme. For instance, in class 1, \textsf{FedSat} achieves an accuracy of 76\%, while the best-performing baseline reaches only 66\%, and the worst-performing baseline achieves a mere 35\%. Unlike the baselines, \textsf{FedSat} performance is consistent across all classes, with maintaining an average accuracy improvement of 28.75\% over the best baseline and 82.5\% over the worst baseline. Also, observe that FedDyn shows slightly higher minimum accuracy but still falls short of SCAFFOLD and \textsf{FedSat}. Whereas Elastic, FedAvg, FedAvgM, and FedProx exhibit low minimum accuracy and inconsistent class-wise accuracy. Additionally, FedLADA demonstrates inconsistent minimum accuracy and highly varying class-wise accuracy.

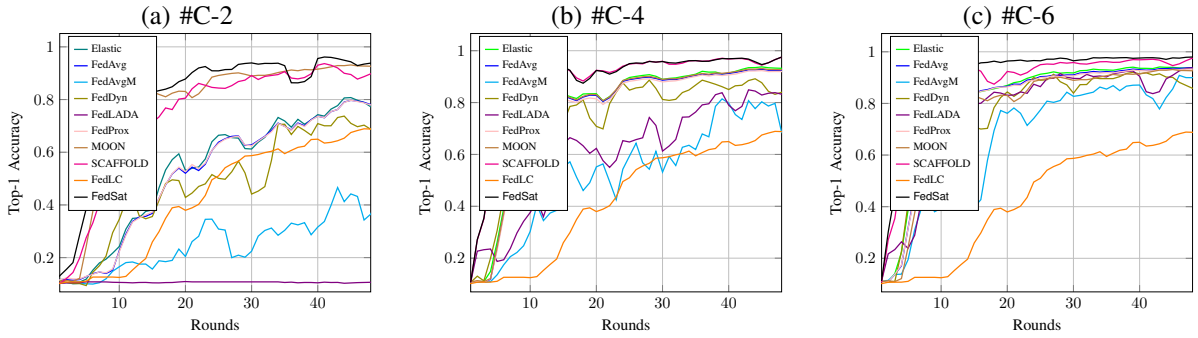
\begin{figure*}[t]
	\centering
	\begin{tabular}{ccc}
		(a) \#C-2 & (b) \#C-4 & (c) \#C-6\\
		\scalebox{0.6}{
			\begin{tikzpicture}
				\begin{axis}[
					xlabel={Rounds},
					ylabel={Top-1 Accuracy},
					grid=both,
					ymin=0.07,
					xmin=1,
					xmax=48,
					legend style={
						at={(0.17,0.99)},
						anchor=north,
						legend cell align=left,
						legend columns=1,
						font=\scriptsize,
						draw=black, 
						fill=white, 
						inner sep=3pt, 
						legend image post style={scale=0.5}, 
					},
					]
					
					\addplot[thick, color=teal] table {Tables/M_lenet5_mnist_nc_2_synthetic_L_CE_lr_0_01_B_32_C_15_E_5_200/elastic.table};
					\addplot[thick, color=blue] table {Tables/M_lenet5_mnist_nc_2_synthetic_L_CE_lr_0_01_B_32_C_15_E_5_200/fedavg.table};
					\addplot[thick, color=cyan] table {Tables/M_lenet5_mnist_nc_2_synthetic_L_CE_lr_0_01_B_32_C_15_E_5_200/fedavgm.table};
					\addplot[thick, color=olive] table {Tables/M_lenet5_mnist_nc_2_synthetic_L_CE_lr_0_01_B_32_C_15_E_5_200/feddyn.table};
					\addplot[thick, color=violet] table {Tables/M_lenet5_mnist_nc_2_synthetic_L_CE_lr_0_01_B_32_C_15_E_5_200/fedlada.table};
					\addplot[thick, color=pink] table {Tables/M_lenet5_mnist_nc_2_synthetic_L_CE_lr_0_01_B_32_C_15_E_5_200/fedprox.table};
					\addplot[thick, color=brown] table
					{Tables/s15_M_lenet5_mnist_nc_4_synthetic_L_CE_lr_0_01_B_16_C_15_E_5_50/moon.table};
					\addplot[thick, color=magenta] table {Tables/M_lenet5_mnist_nc_2_synthetic_L_CE_lr_0_01_B_32_C_15_E_5_200/scaffold.table};
					\addplot[thick, color=orange] table {Tables/M_lenet5_mnist_nc_2_synthetic_L_CE_lr_0_01_B_32_C_15_E_5_200/fedlc.table};
					\addplot[thick, color=black] table {Tables/M_lenet5_mnist_nc_2_synthetic_L_CE_lr_0_01_B_32_C_15_E_5_200/proposedCSS.table};
					
					\addlegendentry{Elastic}
					\addlegendentry {FedAvg}
					\addlegendentry {FedAvgM}
					\addlegendentry {FedDyn}
					\addlegendentry {FedLADA}
					\addlegendentry {FedProx}
					\addlegendentry{MOON}
					\addlegendentry {SCAFFOLD}
					\addlegendentry {FedLC}
					\addlegendentry {\textsf{FedSat}}
					
				\end{axis}
				
			\end{tikzpicture}
		}
		& 
		\scalebox{0.6}{
			\begin{tikzpicture}
				\begin{axis}[
					xlabel={Rounds},
					ylabel={Top-1 Accuracy},
					grid=both,
					ymin=0.07,
					xmin=1,
					xmax=48,
					legend style={
						at={(0.17,0.99)},
						anchor=north,
						legend cell align=left,
						legend columns=1,
						font=\scriptsize,
						draw=black, 
						fill=white, 
						inner sep=3pt, 
						legend image post style={scale=0.5}, 
					},
					]
					
					\addplot[thick, color=green] table {Tables/s15_M_lenet5_mnist_nc_4_synthetic_L_CE_lr_0_01_B_16_C_15_E_5_50/elastic.table};
					\addplot[thick, color=blue] table {Tables/s15_M_lenet5_mnist_nc_4_synthetic_L_CE_lr_0_01_B_16_C_15_E_5_50/fedavg.table};
					\addplot[thick, color=cyan] table {Tables/s15_M_lenet5_mnist_nc_4_synthetic_L_CE_lr_0_01_B_16_C_15_E_5_50/fedavgm.table};
					\addplot[thick, color=olive] table {Tables/s15_M_lenet5_mnist_nc_4_synthetic_L_CE_lr_0_01_B_16_C_15_E_5_50/feddyn.table};
					\addplot[thick, color=violet] table {Tables/s15_M_lenet5_mnist_nc_4_synthetic_L_CE_lr_0_01_B_16_C_15_E_5_50/fedlada.table};
					\addplot[thick, color=pink] table {Tables/s15_M_lenet5_mnist_nc_4_synthetic_L_CE_lr_0_01_B_16_C_15_E_5_50/fedprox.table};
					\addplot[thick, color=brown] table
					{Tables/s15_M_lenet5_mnist_nc_4_synthetic_L_CE_lr_0_01_B_16_C_15_E_5_50/moon.table};
					\addplot[thick, color=magenta] table {Tables/s15_M_lenet5_mnist_nc_4_synthetic_L_CE_lr_0_01_B_16_C_15_E_5_50/scaffold.table};
					\addplot[thick, color=orange] table {Tables/M_lenet5_mnist_nc_2_synthetic_L_CE_lr_0_01_B_32_C_15_E_5_200/fedlc.table};
					\addplot[thick, color=black] table {Tables/s15_M_lenet5_mnist_nc_4_synthetic_L_CE_lr_0_01_B_16_C_15_E_5_50/fedproposedScaffold.table};
					
					\addlegendentry{Elastic}
					\addlegendentry {FedAvg}
					\addlegendentry {FedAvgM}
					\addlegendentry {FedDyn}
					\addlegendentry {FedLADA}
					\addlegendentry {FedProx}
					\addlegendentry{MOON}
					\addlegendentry {SCAFFOLD}
					\addlegendentry {FedLC}
					\addlegendentry {\textsf{FedSat}}
					
				\end{axis}
				
			\end{tikzpicture}
		}
		& 
		\scalebox{0.6}{
			\begin{tikzpicture}
				\begin{axis}[
					xlabel={Rounds},
					ylabel={Top-1 Accuracy},
					grid=both,
					ymin=0.07,
					xmin=1,
					xmax=48,
					legend style={
						at={(0.17,0.99)},
						anchor=north,
						legend cell align=left,
						legend columns=1,
						font=\scriptsize,
						draw=black, 
						fill=white, 
						inner sep=3pt, 
						legend image post style={scale=0.5}, 
					},
					]
					
					\addplot[thick, color=green] table {Tables/s15_M_lenet5_mnist_nc_6_synthetic_L_CE_lr_0_01_B_16_C_15_E_5_50/elastic.table};
					\addplot[thick, color=blue] table {Tables/s15_M_lenet5_mnist_nc_6_synthetic_L_CE_lr_0_01_B_16_C_15_E_5_50/fedavg.table};
					\addplot[thick, color=cyan] table {Tables/s15_M_lenet5_mnist_nc_6_synthetic_L_CE_lr_0_01_B_16_C_15_E_5_50/fedavgm.table};
					\addplot[thick, color=olive] table {Tables/s15_M_lenet5_mnist_nc_6_synthetic_L_CE_lr_0_01_B_16_C_15_E_5_50/feddyn.table};
					\addplot[thick, color=violet] table {Tables/s15_M_lenet5_mnist_nc_6_synthetic_L_CE_lr_0_01_B_16_C_15_E_5_50/fedlada.table};
					\addplot[thick, color=pink] table {Tables/s15_M_lenet5_mnist_nc_6_synthetic_L_CE_lr_0_01_B_16_C_15_E_5_50/fedprox.table};
					\addplot[thick, color=brown] table
					{Tables/s15_M_lenet5_mnist_nc_4_synthetic_L_CE_lr_0_01_B_16_C_15_E_5_50/moon.table};
					\addplot[thick, color=magenta] table {Tables/s15_M_lenet5_mnist_nc_4_synthetic_L_CE_lr_0_01_B_16_C_15_E_5_50/fedproposedScaffold.table};
					\addplot[thick, color=orange] table {Tables/M_lenet5_mnist_nc_2_synthetic_L_CE_lr_0_01_B_32_C_15_E_5_200/fedlc.table};
					\addplot[thick, color=black] table {Tables/s15_M_lenet5_mnist_nc_6_synthetic_L_CE_lr_0_01_B_16_C_15_E_5_50/scaffold.table};
					
					\addlegendentry{Elastic}
					\addlegendentry {FedAvg}
					\addlegendentry {FedAvgM}
					\addlegendentry {FedDyn}
					\addlegendentry {FedLADA}
					\addlegendentry {FedProx}
					\addlegendentry{MOON}
					\addlegendentry {SCAFFOLD}
					\addlegendentry {FedLC}
					\addlegendentry {\textsf{FedSat}}
					
				\end{axis}
				
			\end{tikzpicture}
		} \\
	\end{tabular}
	\caption{Top1-accuracy on MNIST Dataset using LeNet5 model architecture.}
	\label{fig:mnist_eff}
\end{figure*}

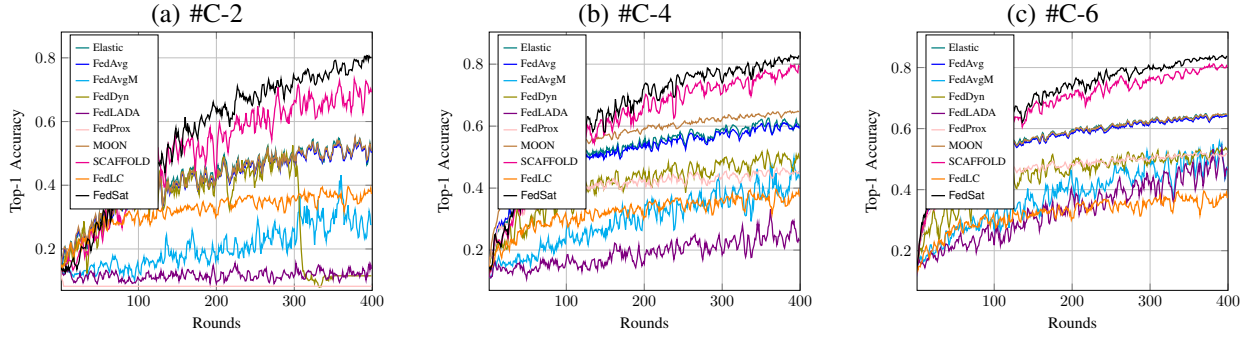
\begin{figure*}[t]
	\centering
	\begin{tabular}{ccc}
		(a) \#C-2 & (b) \#C-4 & (c) \#C-6\\
		\scalebox{0.6}{
			\begin{tikzpicture}
				\begin{axis}[
					xlabel={Rounds},
					ylabel={Top-1 Accuracy},
					grid=both,
					ymin=0.07,
					xmin=1,
					xmax=400,
					legend style={
						at={(0.17,0.99)},
						anchor=north,
						legend cell align=left,
						legend columns=1,
						font=\scriptsize,
						draw=black, 
						fill=white, 
						inner sep=3pt, 
						legend image post style={scale=0.5}, 
					},
					]
					
					\addplot[thick, color=teal] table {Tables/M_resnet8_cifar_nc_2_synthetic_L_CE_lr_0_01_B_16_C_15_E_5_400/elastic.table};
					\addplot[thick, color=blue] table {Tables/M_resnet8_cifar_nc_2_synthetic_L_CE_lr_0_01_B_16_C_15_E_5_400/fedavg.table};
					\addplot[thick, color=cyan] table {Tables/M_resnet8_cifar_nc_2_synthetic_L_CE_lr_0_01_B_16_C_15_E_5_400/fedavgm.table};
					\addplot[thick, color=olive] table {Tables/M_resnet8_cifar_nc_2_synthetic_L_CE_lr_0_01_B_16_C_15_E_5_400/feddyn.table};
					\addplot[thick, color=violet] table {Tables/M_resnet8_cifar_nc_2_synthetic_L_CE_lr_0_01_B_16_C_15_E_5_400/fedlada.table};
					\addplot[thick, color=pink] table {Tables/M_resnet8_cifar_nc_2_synthetic_L_CE_lr_0_01_B_16_C_15_E_5_400/fedprox.table};
					\addplot[thick, color=brown] table
					{Tables/M_resnet8_cifar_nc_2_synthetic_L_CE_lr_0_01_B_16_C_15_E_5_400/moon.table};
					\addplot[thick, color=magenta] table {Tables/M_resnet8_cifar_nc_2_synthetic_L_CE_lr_0_01_B_16_C_15_E_5_400/scaffold.table};
					\addplot[thick, color=orange] table {Tables/M_resnet8_cifar_nc_2_synthetic_L_CE_lr_0_01_B_16_C_15_E_5_400/fedlc.table};
					\addplot[thick, color=black] table {Tables/M_resnet8_cifar_nc_2_synthetic_L_CE_lr_0_01_B_16_C_15_E_5_400/proposedCSS.table};
					
					\addlegendentry{Elastic}
					\addlegendentry {FedAvg}
					\addlegendentry {FedAvgM}
					\addlegendentry {FedDyn}
					\addlegendentry {FedLADA}
					\addlegendentry {FedProx}
					\addlegendentry{MOON}
					\addlegendentry {SCAFFOLD}
					\addlegendentry {FedLC}
					\addlegendentry {\textsf{FedSat}}
					
				\end{axis}
				
			\end{tikzpicture}
		}
		& 
		\scalebox{0.6}{
			\begin{tikzpicture}
				\begin{axis}[
					xlabel={Rounds},
					ylabel={Top-1 Accuracy},
					grid=both,
					ymin=0.07,
					xmin=1,
					xmax=400,
					legend style={
						at={(0.17,0.99)},
						anchor=north,
						legend cell align=left,
						legend columns=1,
						font=\scriptsize,
						draw=black, 
						fill=white, 
						inner sep=3pt, 
						legend image post style={scale=0.5}, 
					},
					]
					
					\addplot[thick, color=teal] table {Tables/M_resnet8_cifar_nc_4_synthetic_L_CE_lr_0_01_B_16_C_15_E_5_400/elastic.table};
					\addplot[thick, color=blue] table {Tables/M_resnet8_cifar_nc_4_synthetic_L_CE_lr_0_01_B_16_C_15_E_5_400/fedavg.table};
					\addplot[thick, color=cyan] table {Tables/M_resnet8_cifar_nc_4_synthetic_L_CE_lr_0_01_B_16_C_15_E_5_400/fedavgm.table};
					\addplot[thick, color=olive] table {Tables/M_resnet8_cifar_nc_4_synthetic_L_CE_lr_0_01_B_16_C_15_E_5_400/feddyn.table};
					\addplot[thick, color=violet] table {Tables/M_resnet8_cifar_nc_4_synthetic_L_CE_lr_0_01_B_16_C_15_E_5_400/fedlada.table};
					\addplot[thick, color=pink] table {Tables/M_resnet8_cifar_nc_4_synthetic_L_CE_lr_0_01_B_16_C_15_E_5_400/fedprox.table};
					\addplot[thick, color=brown] table
					{Tables/M_resnet8_cifar_nc_6_synthetic_L_CE_lr_0_01_B_16_C_15_E_5_400/moon.table};
					\addplot[thick, color=magenta] table {Tables/M_resnet8_cifar_nc_4_synthetic_L_CE_lr_0_01_B_16_C_15_E_5_400/scaffold.table};
					\addplot[thick, color=orange] table {Tables/M_resnet8_cifar_nc_2_synthetic_L_CE_lr_0_01_B_16_C_15_E_5_400/fedlc.table};
					\addplot[thick, color=black] table {Tables/M_resnet8_cifar_nc_4_synthetic_L_CE_lr_0_01_B_16_C_15_E_5_400/proposedCSS.table};
					
					\addlegendentry{Elastic}
					\addlegendentry {FedAvg}
					\addlegendentry {FedAvgM}
					\addlegendentry {FedDyn}
					\addlegendentry {FedLADA}
					\addlegendentry {FedProx}
					\addlegendentry{MOON}
					\addlegendentry {SCAFFOLD}
					\addlegendentry {FedLC}
					\addlegendentry {\textsf{FedSat}}
					
				\end{axis}
				
			\end{tikzpicture}
		}
		& 
		\scalebox{0.6}{
			\begin{tikzpicture}
				\begin{axis}[
					xlabel={Rounds},
					ylabel={Top-1 Accuracy},
					grid=both,
					ymin=0.07,
					xmin=1,
					xmax=400,
					legend style={
						at={(0.17,0.99)},
						anchor=north,
						legend cell align=left,
						legend columns=1,
						font=\scriptsize,
						draw=black, 
						fill=white, 
						inner sep=3pt, 
						legend image post style={scale=0.5}, 
					},
					]
					
					\addplot[thick, color=teal] table {Tables/M_resnet8_cifar_nc_6_synthetic_L_CE_lr_0_01_B_16_C_15_E_5_400/elastic.table};
					\addplot[thick, color=blue] table {Tables/M_resnet8_cifar_nc_6_synthetic_L_CE_lr_0_01_B_16_C_15_E_5_400/fedavg.table};
					\addplot[thick, color=cyan] table {Tables/M_resnet8_cifar_nc_6_synthetic_L_CE_lr_0_01_B_16_C_15_E_5_400/fedavgm.table};
					\addplot[thick, color=olive] table {Tables/M_resnet8_cifar_nc_6_synthetic_L_CE_lr_0_01_B_16_C_15_E_5_400/feddyn.table};
					\addplot[thick, color=violet] table {Tables/M_resnet8_cifar_nc_6_synthetic_L_CE_lr_0_01_B_16_C_15_E_5_400/fedlada.table};
					\addplot[thick, color=pink] table {Tables/M_resnet8_cifar_nc_6_synthetic_L_CE_lr_0_01_B_16_C_15_E_5_400/fedprox.table};
					\addplot[thick, color=brown] table
					{Tables/M_resnet8_cifar_nc_6_synthetic_L_CE_lr_0_01_B_16_C_15_E_5_400/moon.table};
					\addplot[thick, color=magenta] table {Tables/M_resnet8_cifar_nc_6_synthetic_L_CE_lr_0_01_B_16_C_15_E_5_400/scaffold.table};
					\addplot[thick, color=orange] table {Tables/M_resnet8_cifar_nc_2_synthetic_L_CE_lr_0_01_B_16_C_15_E_5_400/fedlc.table};
					\addplot[thick, color=black] table {Tables/M_resnet8_cifar_nc_6_synthetic_L_CE_lr_0_01_B_16_C_15_E_5_400/proposedCSS.table};
					
					\addlegendentry{Elastic}
					\addlegendentry {FedAvg}
					\addlegendentry {FedAvgM}
					\addlegendentry {FedDyn}
					\addlegendentry {FedLADA}
					\addlegendentry {FedProx}
					\addlegendentry {MOON}
					\addlegendentry {SCAFFOLD}
					\addlegendentry {FedLC}
					\addlegendentry {\textsf{FedSat}}
					
				\end{axis}
				
			\end{tikzpicture}
		}  \\
	\end{tabular}
	\caption{Top-1 accuracy on CIFAR10 dataset using ResNet8 model architecture.}
	\label{fig:cifar10_eff}
\end{figure*}

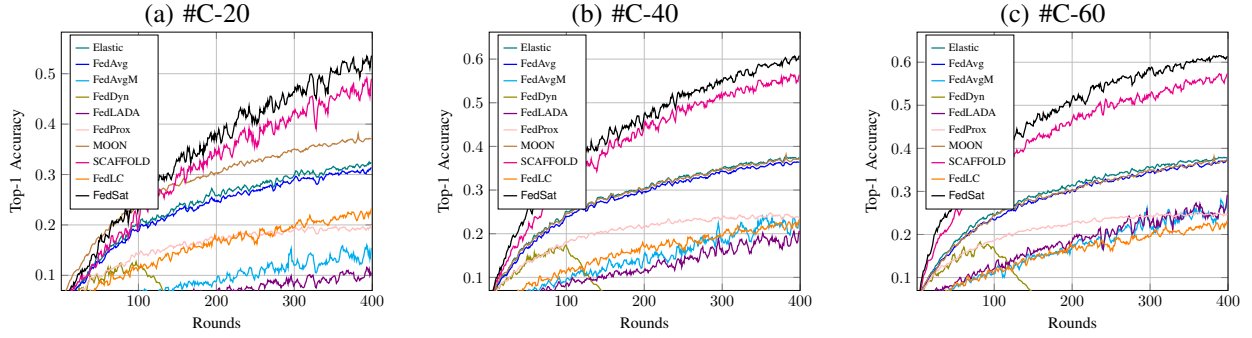
\begin{figure*}[t]
	\centering
	\begin{tabular}{ccc}
		(a) \#C-20 & (b) \#C-40 & (c) \#C-60\\
		\scalebox{0.6}{
			\begin{tikzpicture}
				\begin{axis}[
					xlabel={Rounds},
					ylabel={Top-1 Accuracy},
					grid=both,
					ymin=0.07,
					xmin=1,
					xmax=400,
					legend style={
						at={(0.17,0.99)},
						anchor=north,
						legend cell align=left,
						legend columns=1,
						font=\scriptsize,
						draw=black, 
						fill=white, 
						inner sep=3pt, 
						legend image post style={scale=0.5}, 
					},
					]
					
					\addplot[thick, color=teal] table {Tables/M_resnet8_cifar100_nc_20_synthetic_L_CE_lr_0_01_B_16_C_15_E_5_400/elastic.table};
					\addplot[thick, color=blue] table {Tables/M_resnet8_cifar100_nc_20_synthetic_L_CE_lr_0_01_B_16_C_15_E_5_400/fedavg.table};
					\addplot[thick, color=cyan] table {Tables/M_resnet8_cifar100_nc_20_synthetic_L_CE_lr_0_01_B_16_C_15_E_5_400/fedavgm.table};
					\addplot[thick, color=olive] table {Tables/M_resnet8_cifar100_nc_20_synthetic_L_CE_lr_0_01_B_16_C_15_E_5_400/feddyn.table};
					\addplot[thick, color=violet] table {Tables/M_resnet8_cifar100_nc_20_synthetic_L_CE_lr_0_01_B_16_C_15_E_5_400/fedlada.table};
					\addplot[thick, color=pink] table {Tables/M_resnet8_cifar100_nc_20_synthetic_L_CE_lr_0_01_B_16_C_15_E_5_400/fedprox.table};
					\addplot[thick, color=brown] table
					{Tables/M_resnet8_cifar100_nc_40_synthetic_L_CE_lr_0_01_B_16_C_15_E_5_400/moon.table};
					\addplot[thick, color=magenta] table {Tables/M_resnet8_cifar100_nc_20_synthetic_L_CE_lr_0_01_B_16_C_15_E_5_400/scaffold.table};
					\addplot[thick, color=orange] table {Tables/M_resnet8_cifar100_nc_20_synthetic_L_CE_lr_0_01_B_16_C_15_E_5_400/fedlc.table};
					\addplot[thick, color=black] table {Tables/M_resnet8_cifar100_nc_20_synthetic_L_CE_lr_0_01_B_16_C_15_E_5_400/proposedCSS.table};
					
					\addlegendentry{Elastic}
					\addlegendentry {FedAvg}
					\addlegendentry {FedAvgM}
					\addlegendentry {FedDyn}
					\addlegendentry {FedLADA}
					\addlegendentry {FedProx}
					\addlegendentry{MOON}
					\addlegendentry {SCAFFOLD}
					\addlegendentry {FedLC}
					\addlegendentry {\textsf{FedSat}}
					
				\end{axis}
				
			\end{tikzpicture}
		}
		& 
		\scalebox{0.6}{
			\begin{tikzpicture}
				\begin{axis}[
					xlabel={Rounds},
					ylabel={Top-1 Accuracy},
					grid=both,
					ymin=0.07,
					xmin=1,
					xmax=400,
					legend style={
						at={(0.17,0.99)},
						anchor=north,
						legend cell align=left,
						legend columns=1,
						font=\scriptsize,
						draw=black, 
						fill=white, 
						inner sep=3pt, 
						legend image post style={scale=0.5}, 
					},
					]
					
					\addplot[thick, color=teal] table {Tables/M_resnet8_cifar100_nc_40_synthetic_L_CE_lr_0_01_B_16_C_15_E_5_400/elastic.table};
					\addplot[thick, color=blue] table {Tables/M_resnet8_cifar100_nc_40_synthetic_L_CE_lr_0_01_B_16_C_15_E_5_400/fedavg.table};
					\addplot[thick, color=cyan] table {Tables/M_resnet8_cifar100_nc_40_synthetic_L_CE_lr_0_01_B_16_C_15_E_5_400/fedavgm.table};
					\addplot[thick, color=olive] table {Tables/M_resnet8_cifar100_nc_40_synthetic_L_CE_lr_0_01_B_16_C_15_E_5_400/feddyn.table};
					\addplot[thick, color=violet] table {Tables/M_resnet8_cifar100_nc_40_synthetic_L_CE_lr_0_01_B_16_C_15_E_5_400/fedlada.table};
					\addplot[thick, color=pink] table {Tables/M_resnet8_cifar100_nc_40_synthetic_L_CE_lr_0_01_B_16_C_15_E_5_400/fedprox.table};
					\addplot[thick, color=brown] table
					{Tables/M_resnet8_cifar100_nc_40_synthetic_L_CE_lr_0_01_B_16_C_15_E_5_400/moon.table};
					\addplot[thick, color=magenta] table {Tables/M_resnet8_cifar100_nc_40_synthetic_L_CE_lr_0_01_B_16_C_15_E_5_400/scaffold.table};
					\addplot[thick, color=orange] table {Tables/M_resnet8_cifar100_nc_20_synthetic_L_CE_lr_0_01_B_16_C_15_E_5_400/fedlc.table};
					\addplot[thick, color=black] table {Tables/M_resnet8_cifar100_nc_40_synthetic_L_CE_lr_0_01_B_16_C_15_E_5_400/proposedCSS.table};
					
					\addlegendentry{Elastic}
					\addlegendentry {FedAvg}
					\addlegendentry {FedAvgM}
					\addlegendentry {FedDyn}
					\addlegendentry {FedLADA}
					\addlegendentry {FedProx}
					\addlegendentry{MOON}
					\addlegendentry {SCAFFOLD}
					\addlegendentry {FedLC}
					\addlegendentry {\textsf{FedSat}}
					
				\end{axis}
				
			\end{tikzpicture}
		}
		& 
		\scalebox{0.6}{
			\begin{tikzpicture}
				\begin{axis}[
					xlabel={Rounds},
					ylabel={Top-1 Accuracy},
					grid=both,
					ymin=0.07,
					xmin=1,
					xmax=400,
					legend style={
						at={(0.17,0.99)},
						anchor=north,
						legend cell align=left,
						legend columns=1,
						font=\scriptsize,
						draw=black, 
						fill=white, 
						inner sep=3pt, 
						legend image post style={scale=0.5}, 
					},
					]
					
					\addplot[thick, color=teal] table {Tables/M_resnet8_cifar100_nc_60_synthetic_L_CE_lr_0_01_B_16_C_15_E_5_400/elastic.table};
					\addplot[thick, color=blue] table {Tables/M_resnet8_cifar100_nc_60_synthetic_L_CE_lr_0_01_B_16_C_15_E_5_400/fedavg.table};
					\addplot[thick, color=cyan] table {Tables/M_resnet8_cifar100_nc_60_synthetic_L_CE_lr_0_01_B_16_C_15_E_5_400/fedavgm.table};
					\addplot[thick, color=olive] table {Tables/M_resnet8_cifar100_nc_60_synthetic_L_CE_lr_0_01_B_16_C_15_E_5_400/feddyn.table};
					\addplot[thick, color=violet] table {Tables/M_resnet8_cifar100_nc_60_synthetic_L_CE_lr_0_01_B_16_C_15_E_5_400/fedlada.table};
					\addplot[thick, color=pink] table {Tables/M_resnet8_cifar100_nc_60_synthetic_L_CE_lr_0_01_B_16_C_15_E_5_400/fedprox.table};
					\addplot[thick, color=brown] table
					{Tables/M_resnet8_cifar100_nc_40_synthetic_L_CE_lr_0_01_B_16_C_15_E_5_400/moon.table};
					\addplot[thick, color=magenta] table {Tables/M_resnet8_cifar100_nc_60_synthetic_L_CE_lr_0_01_B_16_C_15_E_5_400/scaffold.table};
					\addplot[thick, color=orange] table {Tables/M_resnet8_cifar100_nc_20_synthetic_L_CE_lr_0_01_B_16_C_15_E_5_400/fedlc.table};
					\addplot[thick, color=black] table {Tables/M_resnet8_cifar100_nc_60_synthetic_L_CE_lr_0_01_B_16_C_15_E_5_400/proposedCSS.table};
					
					\addlegendentry{Elastic}
					\addlegendentry {FedAvg}
					\addlegendentry {FedAvgM}
					\addlegendentry {FedDyn}
					\addlegendentry {FedLADA}
					\addlegendentry {FedProx}
					\addlegendentry{MOON}
					\addlegendentry {SCAFFOLD}
					\addlegendentry {FedLC}
					\addlegendentry {\textsf{FedSat}}
					
				\end{axis}
				
			\end{tikzpicture}
		} \\
	\end{tabular}
	\caption{Top-1 accuracy on CIFAR-100 dataset using ResNet-8 model architecture.}
	\label{fig:cifar100_eff}
\end{figure*}

\subsubsection{\textbf{Convergence}}
Figures~\ref{fig:mnist_eff},~\ref{fig:cifar10_eff}, and~\ref{fig:cifar100_eff} depict the learning efficiency of \textsf{FedSat} compared with the baselines on the MNIST, CIFAR-10, and CIFAR-100 datasets, respectively. 

Across various settings with different numbers of clients, \textsf{FedSat} consistently learns faster and achieves higher accuracy than the other methods. On MNIST (Figure~\ref{fig:mnist_eff}), \textsf{FedSat} converges rapidly within the first 10-20 rounds, reaching an accuracy close to 100\%, while the baselines require more rounds to achieve similar performance. On CIFAR-10 (Figure~\ref{fig:cifar10_eff}), \textsf{FedSat} maintains a clear advantage over the baselines, converging faster and achieving a higher final accuracy. The performance gap is even more significant on the more challenging CIFAR-100 dataset (Figure~\ref{fig:cifar100_eff}), where \textsf{FedSat} outperforms the baselines by a considerable margin in terms of both convergence speed and final accuracy. These observations highlight \textsf{FedSat}'s effectiveness in handling data heterogeneity and its ability to learn efficiently in various federated settings, which can be attributed to its novel aggregation scheme and prediction-sensitive loss function.

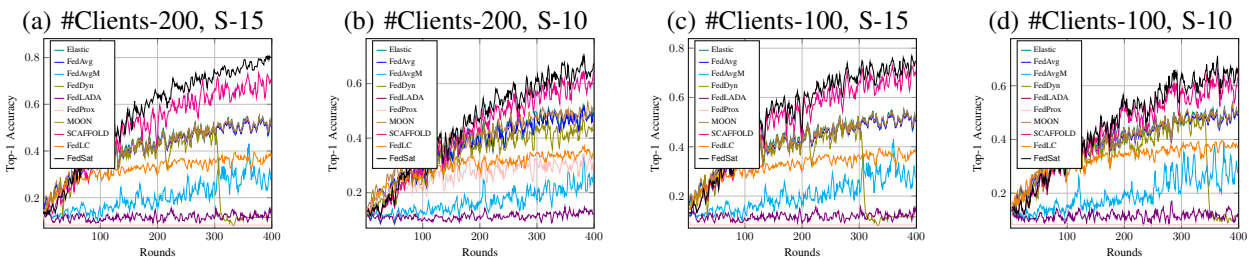
\begin{figure*}[hbp]
	\centering
	\begin{tabular}{cccc}
		(a) \#Clients-200, S-15 & (b)  \#Clients-200, S-10 & (c) \#Clients-100, S-15 & (d)  \#Clients-100, S-10\\
		\scalebox{0.44}{
			\begin{tikzpicture}
				\begin{axis}[
					xlabel={Rounds},
					ylabel={Top-1 Accuracy},
					grid=both,
					ymin=0.07,
					xmin=1,
					xmax=400,
					legend style={
						at={(0.17,0.99)},
						anchor=north,
						legend cell align=left,
						legend columns=1,
						font=\scriptsize,
						draw=black, 
						fill=white, 
						inner sep=3pt, 
						legend image post style={scale=0.5}, 
					},
					]
					
					\addplot[thick, color=teal] table {Tables/M_resnet8_cifar_nc_2_synthetic_L_CE_lr_0_01_B_16_C_15_E_5_400/elastic.table};
					\addplot[thick, color=blue] table {Tables/M_resnet8_cifar_nc_2_synthetic_L_CE_lr_0_01_B_16_C_15_E_5_400/fedavg.table};
					\addplot[thick, color=cyan] table {Tables/M_resnet8_cifar_nc_2_synthetic_L_CE_lr_0_01_B_16_C_15_E_5_400/fedavgm.table};
					\addplot[thick, color=olive] table {Tables/M_resnet8_cifar_nc_2_synthetic_L_CE_lr_0_01_B_16_C_15_E_5_400/feddyn.table};
					\addplot[thick, color=violet] table {Tables/M_resnet8_cifar_nc_2_synthetic_L_CE_lr_0_01_B_16_C_15_E_5_400/fedlada.table};
					\addplot[thick, color=pink] table {Tables/M_resnet8_cifar_nc_2_synthetic_L_CE_lr_0_01_B_16_C_15_E_5_400/fedprox.table};
					\addplot[thick, color=brown] table {Tables/M_resnet8_cifar10_nc_2_synthetic_L_CE_lr_0_01_B_16_C_15_E_5_400/moon.table};
					\addplot[thick, color=magenta] table {Tables/M_resnet8_cifar_nc_2_synthetic_L_CE_lr_0_01_B_16_C_15_E_5_400/scaffold.table};
					\addplot[thick, color=orange] table {Tables/M_resnet8_cifar_nc_2_synthetic_L_CE_lr_0_01_B_16_C_15_E_5_400/fedlc.table};
					\addplot[thick, color=black] table {Tables/M_resnet8_cifar_nc_2_synthetic_L_CE_lr_0_01_B_16_C_15_E_5_400/proposedCSS.table};
					
					\addlegendentry{Elastic}
					\addlegendentry {FedAvg}
					\addlegendentry {FedAvgM}
					\addlegendentry {FedDyn}
					\addlegendentry {FedLADA}
					\addlegendentry {FedProx}
					\addlegendentry{MOON}
					\addlegendentry {SCAFFOLD}
					\addlegendentry {FedLC}
					\addlegendentry {\textsf{FedSat}}
					
				\end{axis}
				
			\end{tikzpicture}
		}
		& 
		\scalebox{0.44}{
			\begin{tikzpicture}
				\begin{axis}[
					xlabel={Rounds},
					ylabel={Top-1 Accuracy},
					grid=both,
					ymin=0.07,
					xmin=1,
					xmax=400,
					legend style={
						at={(0.17,0.99)},
						anchor=north,
						legend cell align=left,
						legend columns=1,
						font=\scriptsize,
						draw=black, 
						fill=white, 
						inner sep=3pt, 
						legend image post style={scale=0.5}, 
					},
					]
					
					\addplot[thick, color=teal] table {Tables/M_resnet8_cifar10_nc_2_synthetic_L_CE_lr_0_01_B_16_C_210_E_5_400/elastic.table};
					\addplot[thick, color=blue] table {Tables/M_resnet8_cifar10_nc_2_synthetic_L_CE_lr_0_01_B_16_C_210_E_5_400/fedavg.table};
					\addplot[thick, color=cyan] table {Tables/M_resnet8_cifar10_nc_2_synthetic_L_CE_lr_0_01_B_16_C_210_E_5_400/fedavgm.table};
					\addplot[thick, color=olive] table {Tables/M_resnet8_cifar10_nc_2_synthetic_L_CE_lr_0_01_B_16_C_210_E_5_400/feddyn.table};
					\addplot[thick, color=violet] table {Tables/M_resnet8_cifar10_nc_2_synthetic_L_CE_lr_0_01_B_16_C_210_E_5_400/fedlada.table};
					\addplot[thick, color=pink] table {Tables/M_resnet8_cifar10_nc_2_synthetic_L_CE_lr_0_01_B_16_C_210_E_5_400/fedprox.table};
					\addplot[thick, color=brown] table {Tables/M_resnet8_cifar10_nc_2_synthetic_L_CE_lr_0_01_B_16_C_10_E_5_400/moon.table};
					\addplot[thick, color=magenta] table {Tables/M_resnet8_cifar10_nc_2_synthetic_L_CE_lr_0_01_B_16_C_210_E_5_400/scaffold.table};
					\addplot[thick, color=orange] table {Tables/M_resnet8_cifar10_nc_2_synthetic_L_CE_lr_0_01_B_16_C_210_E_5_400/fedlc.table};
					\addplot[thick, color=black] table {Tables/M_resnet8_cifar10_nc_2_synthetic_L_CE_lr_0_01_B_16_C_210_E_5_400/proposedCS.table};
					
					\addlegendentry{Elastic}
					\addlegendentry {FedAvg}
					\addlegendentry {FedAvgM}
					\addlegendentry {FedDyn}
					\addlegendentry {FedLADA}
					\addlegendentry {FedProx}
					\addlegendentry{MOON}
					\addlegendentry {SCAFFOLD}
					\addlegendentry {FedLC}
					\addlegendentry {\textsf{FedSat}}
					
				\end{axis}
				
			\end{tikzpicture}
		}
		& 
		\scalebox{0.44}{
			\begin{tikzpicture}
				\begin{axis}[
					xlabel={Rounds},
					ylabel={Top-1 Accuracy},
					grid=both,
					ymin=0.07,
					xmin=1,
					xmax=400,
					legend style={
						at={(0.17,0.99)},
						anchor=north,
						legend cell align=left,
						legend columns=1,
						font=\scriptsize,
						draw=black, 
						fill=white, 
						inner sep=3pt, 
						legend image post style={scale=0.5}, 
					},
					]
					
					\addplot[thick, color=teal] table {Tables/M_resnet8_cifar10_nc_2_synthetic_L_CE_lr_0_01_B_16_C_15_E_5_400/elastic.table};
					\addplot[thick, color=blue] table {Tables/M_resnet8_cifar10_nc_2_synthetic_L_CE_lr_0_01_B_16_C_15_E_5_400/fedavg.table};
					\addplot[thick, color=cyan] table {Tables/M_resnet8_cifar10_nc_2_synthetic_L_CE_lr_0_01_B_16_C_15_E_5_400/fedavgm.table};
					\addplot[thick, color=olive] table {Tables/M_resnet8_cifar10_nc_2_synthetic_L_CE_lr_0_01_B_16_C_15_E_5_400/feddyn.table};
					\addplot[thick, color=violet] table {Tables/M_resnet8_cifar10_nc_2_synthetic_L_CE_lr_0_01_B_16_C_15_E_5_400/fedlada.table};
					\addplot[thick, color=pink] table {Tables/M_resnet8_cifar10_nc_2_synthetic_L_CE_lr_0_01_B_16_C_15_E_5_400/fedprox.table};
					\addplot[thick, color=brown] table {Tables/M_resnet8_cifar10_nc_2_synthetic_L_CE_lr_0_01_B_16_C_15_E_5_400/moon.table};
					\addplot[thick, color=magenta] table {Tables/M_resnet8_cifar10_nc_2_synthetic_L_CE_lr_0_01_B_16_C_15_E_5_400/scaffold.table};
					\addplot[thick, color=orange] table {Tables/M_resnet8_cifar10_nc_2_synthetic_L_CE_lr_0_01_B_16_C_15_E_5_400/fedlc.table};
					\addplot[thick, color=black] table {Tables/M_resnet8_cifar10_nc_2_synthetic_L_CE_lr_0_01_B_16_C_15_E_5_400/proposedCS.table};
					
					\addlegendentry{Elastic}
					\addlegendentry {FedAvg}
					\addlegendentry {FedAvgM}
					\addlegendentry {FedDyn}
					\addlegendentry {FedLADA}
					\addlegendentry {FedProx}
					\addlegendentry{MOON}
					\addlegendentry {SCAFFOLD}
					\addlegendentry {FedLC}
					\addlegendentry {\textsf{FedSat}}
					
				\end{axis}
				
			\end{tikzpicture}
		}  
		&
		\scalebox{0.44}{
			\begin{tikzpicture}
				\begin{axis}[
					xlabel={Rounds},
					ylabel={Top-1 Accuracy},
					grid=both,
					ymin=0.07,
					xmin=1,
					xmax=400,
					legend style={
						at={(0.17,0.99)},
						anchor=north,
						legend cell align=left,
						legend columns=1,
						font=\scriptsize,
						draw=black, 
						fill=white, 
						inner sep=3pt, 
						legend image post style={scale=0.5}, 
					},
					]
					
					\addplot[thick, color=teal] table {Tables/M_resnet8_cifar10_nc_2_synthetic_L_CE_lr_0_01_B_16_C_10_E_5_400/elastic.table};
					\addplot[thick, color=blue] table {Tables/M_resnet8_cifar10_nc_2_synthetic_L_CE_lr_0_01_B_16_C_10_E_5_400/fedavg.table};
					\addplot[thick, color=cyan] table {Tables/M_resnet8_cifar10_nc_2_synthetic_L_CE_lr_0_01_B_16_C_10_E_5_400/fedavgm.table};
					\addplot[thick, color=olive] table {Tables/M_resnet8_cifar10_nc_2_synthetic_L_CE_lr_0_01_B_16_C_10_E_5_400/feddyn.table};
					\addplot[thick, color=violet] table {Tables/M_resnet8_cifar10_nc_2_synthetic_L_CE_lr_0_01_B_16_C_10_E_5_400/fedlada.table};
					\addplot[thick, color=pink] table {Tables/M_resnet8_cifar10_nc_2_synthetic_L_CE_lr_0_01_B_16_C_10_E_5_400/fedprox.table};
					\addplot[thick, color=brown] table {Tables/M_resnet8_cifar10_nc_2_synthetic_L_CE_lr_0_01_B_16_C_10_E_5_400/moon.table};
					\addplot[thick, color=magenta] table {Tables/M_resnet8_cifar10_nc_2_synthetic_L_CE_lr_0_01_B_16_C_10_E_5_400/scaffold.table};
					\addplot[thick, color=orange] table {Tables/M_resnet8_cifar10_nc_2_synthetic_L_CE_lr_0_01_B_16_C_10_E_5_400/fedlc.table};
					\addplot[thick, color=black] table {Tables/M_resnet8_cifar10_nc_2_synthetic_L_CE_lr_0_01_B_16_C_10_E_5_400/proposedCS.table};
					
					\addlegendentry{Elastic}
					\addlegendentry {FedAvg}
					\addlegendentry {FedAvgM}
					\addlegendentry {FedDyn}
					\addlegendentry {FedLADA}
					\addlegendentry {FedProx}
					\addlegendentry{MOON}
					\addlegendentry {SCAFFOLD}
					\addlegendentry {FedLC}
					\addlegendentry {\textsf{FedSat}}
					
				\end{axis}
				
			\end{tikzpicture}
		}
		
		\\
	\end{tabular}
	\caption{Comparison of scalibility with baselines on CIFAR10 dataset using ResNet-8 model architecture.}
	\label{fig:cifar10_scalability}
\end{figure*}

\begin{table*}[hbp]
	\centering
	\caption{Performance of \textsf{FedSat} with varying number of workers (w) for evaluation clients updated parameters}
	\label{tab:fedsat_workers}
	\begin{tabular}{lccc}
		\toprule
		\textbf{Dataset} & \multicolumn{3}{c}{\textbf{Number of Workers (w)}} \\
		\cmidrule(lr){2-4}
		& \textbf{w=1}(client itself) & \textbf{w=7} & \textbf{w=15} \\
		\midrule
		\textbf{MNIST}    & 99.29 & 99.34 & 99.47 \\
		\textbf{CIFAR}    & 83.56 & 83.86 & 84.05 \\
		\textbf{CIFAR-100} & 57.85 & 58.36 & 58.79 \\
		\bottomrule
	\end{tabular}
\end{table*}

\begin{table*}[hbp]
	\centering
	\caption{Analysis of \textsf{FedSat's} components}
	\scriptsize
	\label{tab:ablation1}
	\begin{tabular}{lccc}
		\toprule
		Components & MNIST & CIFAR-10 & CIFAR-100 \\
		& (LeNet-5) & (ResNet-8) & (ResNet-18) \\
		\midrule
		Base Algorithm & 97.95 & 86.09 & 66.32 \\		
		Base + PSL & 98.29 & 87.55 & 68.24 \\
		Base + PWA & 98.16 & 87.35 & 67.15 \\
		\textsf{FedSat} (Base + PSL + PWA) & \textbf{98.49} & \textbf{87.67} & \textbf{68.48} \\
		\bottomrule
	\end{tabular}
\end{table*}


\subsubsection{\textbf{Scalability}}
To show the scalability of \textsf{FedSat}, we vary the number of clients on CIFAR-10. Specifically, we consider the following four settings: (1) partitioning the dataset into 200 clients and randomly sampling 15 clients in each round, (2) partitioning the dataset into 200 clients and randomly sampling 10 clients in each round, (3) partitioning the dataset into 100 clients and randomly sampling 15 clients in each round, and (4) partitioning the dataset into 100 clients and randomly sampling 10 clients in each round. It is evident from the Figure~\ref{fig:cifar10_scalability} that across all settings, \textsf{FedSat} maintains its high performance, consistently achieving the highest final accuracy and faster convergence than the baselines. As the number of total clients increases, the performance gap between \textsf{FedSat} and the baselines widens, with \textsf{FedSat} outperforming the best baseline (SCAFFOLD) by 3-12\% in terms of final accuracy. Moreover, \textsf{FedSat}'s performance remains stable and superior to the baselines, regardless of the number of clients selected per round, demonstrating its robustness to variations in client participation. These observations highlight \textsf{FedSat}'s ability to efficiently leverage increased data and computational resources in larger federated networks, making it a promising solution for real-world, large-scale federated learning applications.

Readers may refer to Appendix B in the supplementary document attached with this manuscript for additional experimental results.

\subsection{Impact of Worker-set on FedSat's Performance}

The benefit of having a worker-set for validating client's updated parameters are twofold: (1) it ensures evaluation fairness and robustness by mitigating the impact of data distribution heterogeneity across clients, as in FL settings each client may possess vastly different data distribution and consequently, parameters evaluation by a single client with a suboptimal data distribution could compromise the fairness of the validation process and potentially impede the generation of a robust global model (2) the worker-set mechanism systematically identifies and filters out malicious or poor-quality parameter updates, thereby enhancing the global model's robustness.
It is worth noting that our proposed framework design is flexible in worker-set size configuration, allowing users to adjust the number of workers involved in parameter validation based on their specific requirements. In resource-constrained scenarios where communication costs are critical, user can chose the minimal configuration (w=1), where the corresponding client itself performs statistical evaluation of its model using a local validation dataset, eliminating additional communication overhead. Furthermore, to show the impact of the number of workers in validating performance of each client's updated parameters, we evaluate \textsf{FedSat} on different datasets with varying worker-set size. As depicted in Table~\ref{tab:fedsat_workers}, the result demonstrates that, as the number of workers increases, the performance of \textsf{FedSat} improves across all datasets. Specifically, for the MNIST dataset, the accuracy increases from 99.29\% with single workers (w=1) to 99.47\% with 15 workers. Similarly, on CIFAR-10 and CIFAR-100, there is a noticeable improvement in accuracy as more workers are involved in the evaluation process. These findings suggest that a higher number of workers for validating updates can enhance the overall performance of the model, especially in more complex datasets like CIFAR-10 and CIFAR-100. 

\subsection{Ablation Study}
To evaluate the contribution of each component in FedSat, we conduct extensive ablation experiments across different datasets and model architectures. Table~\ref{tab:ablation1} presents the results with various combinations of our two key components: prediction-sensitive loss (PSL) and prioritized-class based weighted aggregation (PWA). The prediction-sensitive loss alone improves the best accuracy by 0.38-3.07\% across datasets by better handling class imbalance. The prioritized-class based weighted aggregation provides 0.25-1.49\% gain by optimizing update incorporation. When combined, these components show synergistic effects, with FedSat achieving 0.58-3.43\% overall improvement compared to the base algorithm. The results demonstrate that each component addresses distinct aspects of heterogeneity, and their combination is crucial for optimal performance. On CIFAR-100, removing PSL causes the largest performance drop (3.07\%), followed by PAW (1.49\%), indicating importance of PSL for complex datasets. For simpler datasets like MNIST, the components show more balanced contributions, suggesting their relative importance varies with task complexity.

\section{Conclusion}
\label{sec:conclusion}
This paper introduces \textsf{FedSat}, a novel federated learning approach designed to address the challenges posed by heterogeneous data distributions across clients. By incorporating a prediction-sensitive loss function and a prioritized-class based weighted aggregation scheme, \textsf{FedSat} effectively mitigates the negative effects of label skewness, missing classes, and quantity skewness in federated learning environments. Our extensive experimental evaluation across various datasets (MNIST, CIFAR-10, and CIFAR-100) and model architectures (MLP, LeNet-5, ResNet-8, and ResNet-18) demonstrates the superiority of \textsf{FedSat} over state-of-the-art baseline methods. In particular, \textsf{FedSat} consistently achieves higher accuracy, with an average improvement of 1.8\% over the second-best method and up to 19.87\% over the weakest-performing baseline. \textsf{FedSat}'s robustness is particularly evident in extreme non-IID scenarios, where it demonstrates significant improvements in minimum accuracy for a client by up to 36.25\% on average. This highlights \textsf{FedSat}'s ability to enhance the performance of underrepresented classes and clients, leading to a more balanced and equitable learning outcome. Furthermore, \textsf{FedSat} exhibits faster convergence rates and better scalability compared to existing methods, maintaining its performance advantages even as the number of clients increases. This scalability makes \textsf{FedSat} a promising solution for large-scale, real-world federated learning applications. Our future plan aims to incorporate various privacy preserving techniques, such as differential privacy or secure aggregation, to further enhance its utility in privacy-sensitive domains. Further, we plan to extend blockchain support as a way to prevent various adversarial attacks in \textsf{FedSat}. 


	\bibliographystyle{IEEEtran}
	\bibliography{references}
\includepdf[pages=-]{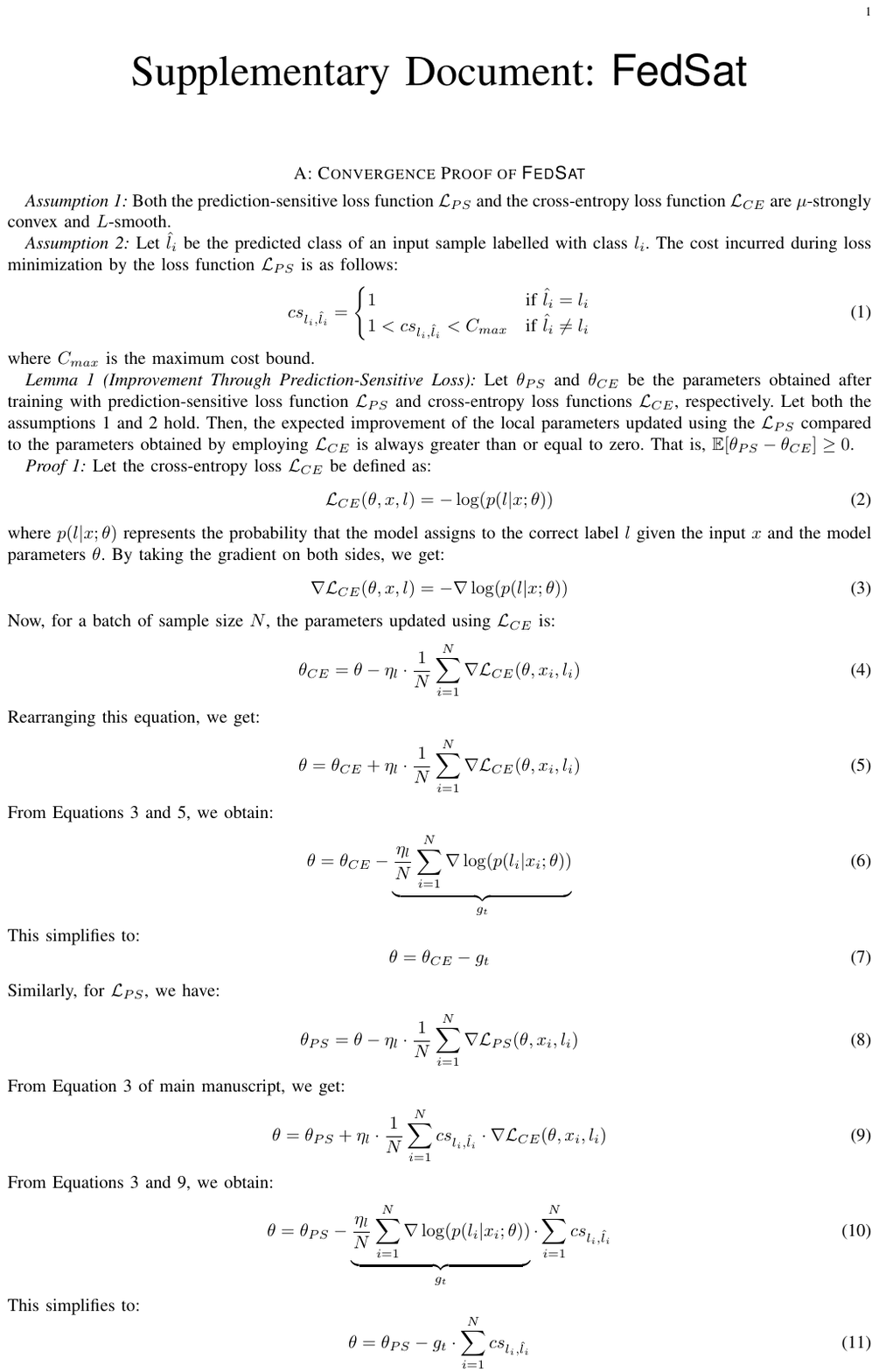}  

\end{document}